%% file: main.tex
\documentclass[10pt,twocolumn,letterpaper]{article}

\usepackage{iccv}
\usepackage{times}
\usepackage{epsfig}
\usepackage{graphicx}
\usepackage{amsmath}
\usepackage{amssymb}

\usepackage{url}

\usepackage{graphicx}
\usepackage{amsmath}
\usepackage{amssymb}

\usepackage{amsmath, amsthm, amssymb}
\usepackage{textcomp}
\usepackage{stmaryrd}
\usepackage{upgreek}
\usepackage{bm}
\usepackage{cases}
\usepackage{mathtools}
\usepackage{arydshln}
\usepackage{multirow}
\usepackage{bbm}
\usepackage{etoc}  
\usepackage{wrapfig}

\usepackage{multirow}
\usepackage{rotating}
\usepackage{booktabs}
\usepackage{tabularx}
\usepackage{tabulary}
\usepackage{rotating}
\usepackage{adjustbox}

\usepackage{caption}

\usepackage{enumitem}

\usepackage[utf8]{inputenc} %
\usepackage{url}            %
\usepackage{booktabs}       %
\usepackage{nicefrac}       %
\usepackage{microtype}      %
\usepackage{xspace}

\usepackage{graphicx}
\usepackage{nccmath} %
\usepackage{algorithm}
\usepackage[noend]{algpseudocode}

\usepackage{algorithm}
\usepackage{listings}

\usepackage[outline]{contour}
\contourlength{0.1pt}
\contournumber{10}%

\usepackage[pagebackref=true,breaklinks=true,letterpaper=true,colorlinks,bookmarks=false]{hyperref}

\iccvfinalcopy %

\def\iccvPaperID{26} %

\ificcvfinal\fi

\usepackage{tikz}
\usepackage{makecell}

\usepackage{amssymb}%
\usepackage{pifont}%
\newcommand{\cmark}{\ding{51}}%
\newcommand{\xmark}{\ding{55}}%

\definecolor{bluebell}{rgb}{0.64, 0.64, 0.82}
\definecolor{cadet}{rgb}{0.33, 0.41, 0.47}
\newcommand{\std}[1]{\tiny \textcolor{bluebell}{$\pm$#1}}
\newcommand{\unfair}[1]{\colorbox{Gray}{#1}}

\usepackage{color, colortbl}
\definecolor{Gray}{gray}{0.9}
\newcommand{\md}[1]{\colorbox{red!15}{#1}}

\usepackage{inconsolata}
\usepackage{subcaption}

\input{utils/macros}

\begin{document}
\title{\pst: Proportional Amplitude Spectrum Training Augmentation for Syn-to-Real Domain Generalization}

\author{
\centerline{Prithvijit Chattopadhyay\thanks{Equal Contribution. Correspondence to \texttt{prithviijt3@gatech.edu}} \quad Kartik Sarangmath$^*$ \quad Vivek Vijaykumar \quad Judy Hoffman}\\
\centerline{Georgia Tech}\\
\centerline{\texttt{\footnotesize \{prithvijit3, ksarangmath3, vivekvjk, judy\}@gatech.edu}} \\
\centerline{\small\tt \url{https://github.com/prithv1/PASTA}}
}

\maketitle
\ificcvfinal\thispagestyle{empty}\fi

\begin{abstract}
    \input{sections/abstract}
\end{abstract}

\section{Introduction}
\input{sections/intro}

\section{Related work}
\input{sections/related_work}

\section{Method}
\input{sections/method}

\section{Experimental Details}
\label{sec:experiments}
\input{sections/experiments_new}

\section{Results and Findings}
\input{sections/results_new}

\section{Conclusion}
\input{sections/conclusion}

\appendix
\section{Appendix}
This appendix is organized as follows. In Sec.~\ref{sec:implementation}, we first expand on implementation and training details from the main paper. Then, in Sec.~\ref{sec:syn2real}, we provide per-class synthetic-to-real generalization results (see Sec. 5.1 of the main paper). Sec.~\ref{sec:pasta_analysis} includes additional discussions surrounding different aspects of \pst. Sec.~\ref{sec:amp_analysis} goes through an empirical analysis of the amplitude 
spectra
for synthetic and real images. Next, Sec.~\ref{sec:qual_examples} contains more qualitative examples of \pst augmentations and predictions for semantic segmentation. Finally, Sec.~\ref{sec:licenses} summarizes the licenses associated with different assets used in our experiments.\looseness=-1

\subsection{Implementation and Training Details}
\label{sec:implementation}
\input{appendix_sections/implementation_details}

\subsection{Synthetic-to-Real Generalization Results}
\label{sec:syn2real}

\input{appendix_sections/syn_to_real_results}

\subsection{\pst Analysis}
\label{sec:pasta_analysis}
\input{appendix_sections/pasta_analysis}

\subsection{Amplitude Analysis}
\label{sec:amp_analysis}
\input{appendix_sections/amplitude_analysis}

\subsection{Qualitative Examples}
\label{sec:qual_examples}
\input{appendix_sections/qualitative_examples}

\subsection{Assets Licenses}
\label{sec:licenses}
\input{appendix_sections/asset_licenses}

{\small
\bibliographystyle{ieee_fullname}
\bibliography{egbib,main}
}

\end{document}

%% file: utils/macros.tex
\newcommand{\pc}{\textcolor{black}}

\newcommand{\ks}{\textcolor{black}}

\newcolumntype{R}[2]{%
    >{\adjustbox{angle=#1,lap=\width-(#2)}\bgroup}%
    l%
    <{\egroup}%
}
\newcommand*\rotb{\multicolumn{1}{R{45}{1em}}}%

\newcommand{\pst}{\textsc{Pasta}\xspace}

\makeatletter
\DeclareRobustCommand\onedot{\futurelet\@let@token\@onedot}
\def\@onedot{\ifx\@let@token.\else.\null\fi\xspace}

\makeatother

\newcommand{\eat}[1]{}
\newcommand{\replace}[2]{} %

\newenvironment{packed_itemize}{
	\begin{list}{\labelitemi}{\leftmargin=1em}
		\vspace{-6pt}
		\setlength{\itemsep}{0pt}
		\setlength{\parskip}{0pt}
		\setlength{\parsep}{0pt}
	}{\end{list}}

%% file: sections/abstract.tex
Synthetic data offers the promise of cheap and bountiful training data for settings where labeled real-world data is scarce. However, models trained on synthetic data significantly underperform 
when evaluated
on real-world data. In this paper, we propose Proportional Amplitude Spectrum Training Augmentation (\pst), a simple and effective augmentation strategy to improve out-of-the-box synthetic-to-real (syn-to-real) generalization performance. \pst  perturbs the amplitude spectra of synthetic images in the Fourier domain to generate augmented views. Specifically, with \pst we propose a structured perturbation strategy where high-frequency components are perturbed relatively more than the low-frequency 
ones.
For the tasks of semantic segmentation (GTAV$\to$Real), object detection (Sim10K$\to$Real), and object recognition (VisDA-C Syn$\to$Real), across 
a total of 
5 
syn-to-real shifts, 
we find that \pst
outperforms more complex state-of-the-art generalization methods while being complementary to the same.\looseness=-1

%% file: sections/intro.tex
For complex tasks, deep models often rely on training with substantial labeled data. 
Real-world data can be expensive to label and 
an available labeled training set often captures only a limited set of real-world appearance diversity. Synthetic data offers an opportunity to cheaply generate diverse samples that can better cover the anticipated variance of real-world test data. 
However, models trained on synthetic data often struggle to generalize to real world data -- e.g., the performance of a vanilla DeepLabv3+~\cite{chen2018encoder} (ResNet-50 backbone) architecture on semantic segmentation drops 
from $73.45$ mIoU on GTAV~\cite{richter2016playing} 
to $28.95$ mIoU
on Cityscapes~\cite{cordts2016cityscapes} for the same set of classes.
Several approaches have been considered
to tackle this problem.

\begin{figure}[t]
\centering
\includegraphics[width=\columnwidth]{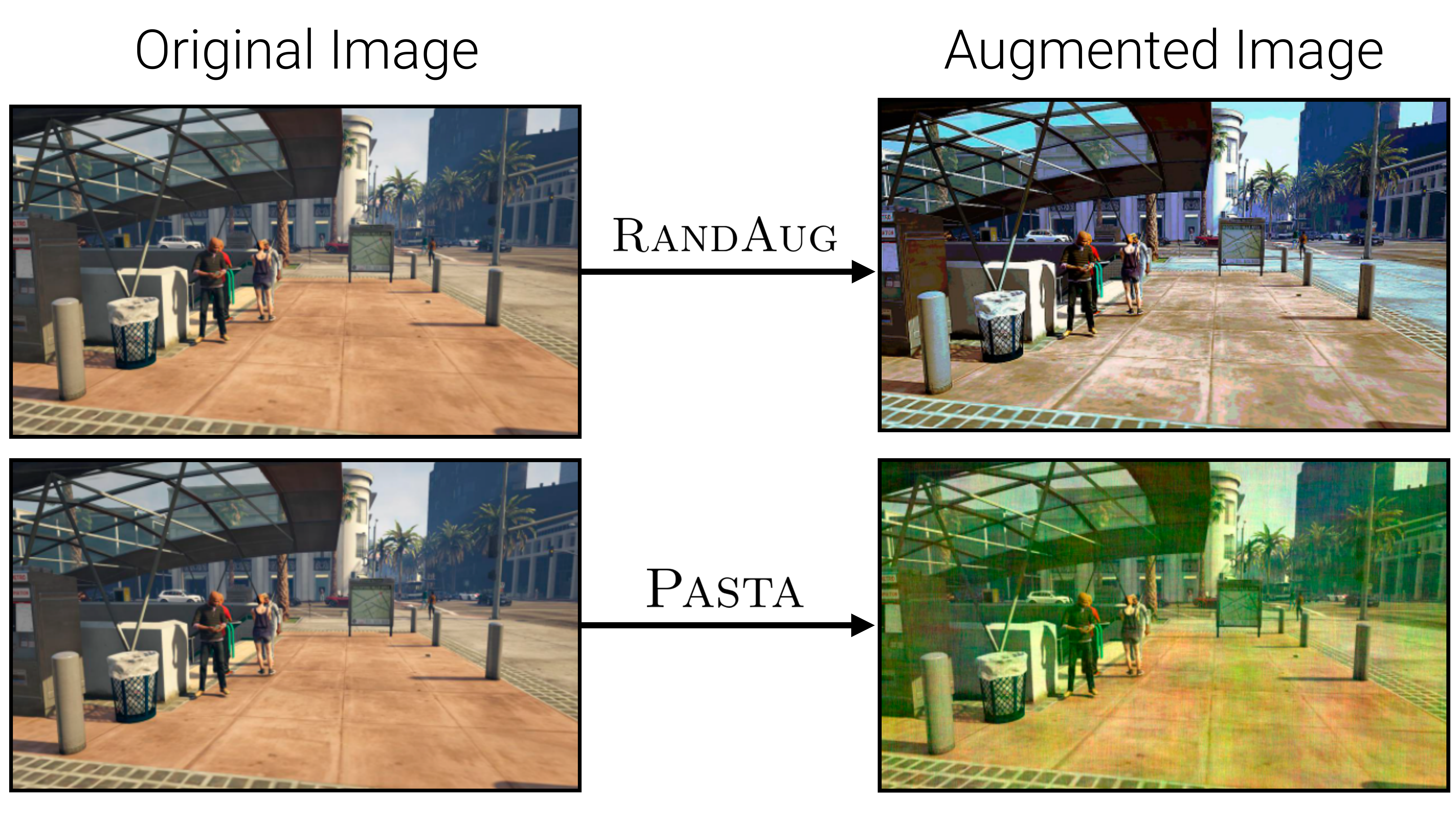}
\vspace{-10pt}
\caption{\small\textbf{\pst augmentation samples.} A synthetic image from GTAV~\cite{richter2016playing} when augmented with RandAugment~\cite{cubuk2020randaugment} \textbf{[Top]} and \pst \textbf{[Bottom]}. We can see that \pst creates augmented views different from existing photometric operations. 
}
\label{fig:teaser}
\vspace{-15pt}
\end{figure}

In this paper, we propose 
a novel
augmentation strategy, called Proportional Amplitude Spectrum Training Augmentation (\pst), 
to address
the synthetic-to-real generalization problem. 
\pst, as an augmentation strategy for synthetic data, aims to
satisfy three key criteria:
(1) \textit{strong out-of-the-box generalization performance}, (2) \textit{plug-and-play} compatibility with existing methods, and (3) benefits across \textit{tasks, backbones,} and \textit{shifts}.
\pst 
achieves this by
perturbing the amplitude spectra (obtained by applying 2D FFT to input images) of the source synthetic images
in the Fourier domain. While prior work 
has explored
augmenting images in the Fourier domain~\cite{xu2021fourier,yang2020fda,huang2021fsdr}, they mostly rely on the observations that -- 
(1) among the amplitude and phase spectra, 
phase
tends to capture more high-level semantics~\cite{1170798,1456290,piotrowski1982demonstration,hansen2007structural,yang2020phase} and (2) low-frequency (LF) bands of the amplitude spectrum tend to capture style information / low-level statistics (illumination, lighting, etc.)~\cite{yang2020fda}.

We further observe
\pc{that synthetic images have less diversity in the high-frequency (HF) bands of their amplitude spectra compared to real images (see Sec.~\ref{sec:pasta} for a detailed discussion).}
Motivated by these key observations, \pst provides a structured way to perturb the amplitude spectra of source synthetic images to ensure that
a model is exposed to more variations in high-frequency components during training. 
We empirically observe that by relying on such a simple set of motivating observations, \pst leads to significant improvements in synthetic-to-real generalization performance -- 
e.g., out-of-the-box GTAV~\cite{richter2016playing}$\to$Cityscapes~\cite{cordts2016cityscapes} generalization performance of a vanilla DeepLabv3+ (ResNet-50 backbone) model
improves 
from $28.95$ mIoU to $44.12$ mIoU -- a $15+$ absolute mIoU point improvement!

\pst involves the following steps. Given an input image, we apply 2D Fast Fourier Transform (FFT) to obtain the corresponding amplitude and phase spectra in the Fourier domain. For every spatial frequency $[m, n]$ in the 
\pc{amplitude}
spectrum, we sample a multiplicative jitter value $\epsilon$ 
such that
the perturbation strength increases monotonically with $[m, n]$,
thereby, ensuring that higher frequency components in the amplitude spectrum are perturbed more compared to the lower frequency ones.
Finally, given the perturbed amplitude and the original phase spectra, we can apply an inverse 2D Fast Fourier Transform (iFFT) to obtain the augmented image.
This simple strategy of 
applying fine-grained structured perturbations to the amplitude spectra of synthetic images
leads to strong out-of-the-box generalization without the need for specialized components, task-specific design, or changes to learning rules.
Fig.~\ref{fig:teaser} shows an example image augmented by \pst.\looseness=-1

To summarize, we make the following contributions:
\begin{packed_itemize}
\item We introduce \textsc{P}roportional \textsc{A}mplitude \textsc{S}pectrum \textsc{T}raining \textsc{A}ugmentation (\pst), a
simple and effective 
augmentation strategy for synthetic-to-real generalization. \pst 
perturbs
the amplitude spectra of synthetic images 
so as to expose a model to more high-frequency variations.
\item We show that applying \pst sets the new state of the art for synthetic-to-real generalization for $3$ tasks -- Semantic Segmentation (GTAV~\cite{richter2016playing}$\to$$\{$Cityscapes~\cite{cordts2016cityscapes}, Mapillary~\cite{neuhold2017mapillary}, BDD100K~\cite{yu2020bdd100k}$\}$), Object Detection (Sim10K~\cite{johnson2016driving}$\to$Cityscapes)
and Object Recognition (VisDA-C~\cite{peng2017visda} Syn$\to$Real) -- covering a total of $5$ syn-to-real shifts 
with
multiple backbones.\looseness=-1
\item Our experimental results demonstrate that \pst 
-- (1) 
frequently enables a baseline model to outperform previous state-of-the-art approaches that rely on specialized architectural components, additional synthetic or real data, or alternate objectives; 
(2) 
is complementary to existing methods; (3) outperforms prior adaptive object detection methods; and (4) either outperforms or is competitive with current augmentation strategies.
\end{packed_itemize}

%% file: sections/related_work.tex
\par \noindent
\textbf{Domain Generalization (DG).} DG typically involves training models on single or multiple labeled data sources to generalize well to novel test time data sources (unseen during training).
Several approaches have been proposed to tackle domain generalization ~\cite{blanchard2011generalizing,muandet2013domain}, such as 
decomposing a model into domain invariant and specific components and utilizing the former to make predictions~\cite{ghifary2015domain,khosla2012undoing}, learning domain specific masks for generalization~\cite{chattopadhyay2020learning}, using meta-learning to train a robust model
~\cite{li2018learning,wang2020meta,balaji2018metareg,chen2022discriminative, dou2019domain}, \ks{manipulating feature statistics to augment training data~\cite{zhou2021domain,li2022ood, nuriel2021adain},} and using models crafted based on risk minimization formalisms~\cite{arjovsky2019invariant}. More recently, properly tuned ERMs (Empirical Risk Minimization) have proven to be a competitive DG approach~\cite{gulrajani2021in}, with follow-up work adopting various optimization and regularization techniques~\cite{shi2022gradient,cha2021swad} on top.

\par \noindent
\textbf{Single Domain Generalization (SDG).} 
Unlike DG which leverages diversity across multiple sources for better generalization, SDG considers generalizing from 
a single source.
Notable approaches for SDG 
use meta-learning~\cite{qiao2020learning} by considering strongly augmented 
source images
as meta-target data (by exposing the model to increasingly distinct augmented views of the source data~\cite{wang2021learning,li2021progressive}) 
and learning 
feature normalization schemes with auxiliary objectives~\cite{fan2021adversarially}.

\par \noindent
\textbf{Synthetic-to-Real Generalization (Syn-to-Real).} 
Prior work on syn-to-real generalization has mostly focused on some specific methods, including learning feature normalization / whitening schemes~\cite{pan2018two,choi2021robustnet}, using external data for style injection~\cite{kim2023wedge,kundu2021generalize}, explicitly optimizing for robustness~\cite{chen2020automated}, leveraging strong augmentations / domain randomization~\cite{yue2019domain,kundu2021generalize}, consistency objectives~\cite{zhao2022shade} and using contrastive techniques to aid generalization~\cite{chen2021contrastive}. Some approaches have also considered adapting from synthetic to real images, using techniques such as adversarial training~\cite{chen2018domain}, adversarial alignment losses~\cite{saito2018adversarial}, balancing transferability and discriminability~\cite{Chen2020HarmonizingTA} and feature alignment~\cite{Vibashan2021MeGACDAMG}.
\pst is more similar to the kind of methods adopting augmentations for improving out-of-the-box generalization
We consider $3$ of the most commonly studied syn-to-real generalization settings 
-- (1) Semantic Segmentation - GTAV~\cite{richter2016playing}$\to$Real,
(2) Object Detection - Sim10K~\cite{johnson2016driving}$\to$Real
and (3) Object Recognition - 
VisDA-C~\cite{peng2017visda} Syn$\rightarrow$Real.~\cite{mishra2021task2sim} recently proposed tailoring synthetic data for better generalization.

\par \noindent
\textbf{Fourier Generalization \& Adaptation Methods.} 
Prior work 
that explored
augmenting images in the 
Fourier
domain (as opposed to the pixel space)
rely on a key
empirical observation~\cite{1170798,1456290,piotrowski1982demonstration,hansen2007structural} that the phase component of the Fourier spectrum 
tends to preserve high-level semantics, and therefore, they 
focused mostly on
perturbing the amplitude. \pst is in line with this style of approach.
Amplitude Jitter (AJ)~\cite{xu2021fourier} and Amplitude Mixup (AM)~\cite{xu2021fourier} 
are methods similar to \pst that augment images by perturbing their amplitude spectra. While AM mixes the amplitude spectra of different images, AJ applies uniform perturbation with a single jitter value $\epsilon$. FSDR~\cite{huang2021fsdr}, on the other hand, isolates domain variant and invariant frequency components using extra data and sets up a learning paradigm.
Building on top of~\cite{xu2021fourier}, ~\cite{yang2021hcdg} adds a significance mask 
when linearly interpolating
amplitudes.~\cite{huang2021rda} 
only perturbs image-frequency components that capture little semantic information.
~\cite{wang2022domain} uses an encoder-decoder to obtain high/low frequency features and augments images by adding noise to high frequency phase and low frequency amplitude.
~\cite{yin2019fourier,chen2021amplitude} study how amplitude and phase perturbations impact robustness to natural corruptions~\cite{hendrycks2019benchmarking}.
In contrast to these works, \pst's simple strategy of perturbing amplitude spectra in a structured way and leads to strong out-of-the-box generalization without the need for specialized components, extra data, task-specific design, or changes to learning rules.

%% file: sections/method.tex
We investigate how well models trained on a single labeled synthetic source dataset generalize to real target data, without any access to target data during training.

\subsection{Preliminaries: Fourier Transform}

\pst creates augmented views by applying perturbations to the Fourier amplitude spectrum. The amplitude $\mathcal{A}(x)$ and phase $\mathcal{P}(x)$ components of Fourier spectra of images have been widely used in image processing for several applications -- for studying properties (e.g., periodic intereferences), compact representations (e.g, JPEG compression), digital filtering, etc -- and more recently for generalizing and adapting deep networks by perturbing the amplitude spectra~\cite{xu2021fourier,yang2020fda}. We now cover preliminaries explaining how to obtain amplitude and phase spectra from an image.

Consider a single-channel image $x\in\mathbb{R}^{H\times W}$. 
The Fourier transform $\mathcal{F}(x)$ of $x$ can be expressed as,
\begin{equation}
    \mathcal{F}(x)[m,n] = \sum_{h=0}^{H-1} \sum_{w=0}^{W-1} x[h,w] \exp\left(-2\pi i\left(\frac{h}{H}m + \frac{w}{W}n\right)\right)
\label{eq:fourier_spec}
\end{equation}

where $i^2 = -1$ and $m,n$ denote spatial frequencies. 

The inverse Fourier transform, $\mathcal{F}^{-1}(\cdot)$, that maps signals from the frequency domain to the image domain can be defined accordingly. Note that the Fourier spectrum $\mathcal{F}(x)\in \mathbb{C}^{H\times W}$. If $\mathbf{Re}(\mathcal{F}(x)[\cdot,\cdot])$ and $\mathbf{Im}(\mathcal{F}(x)[\cdot,\cdot])$ denote the real and imaginary parts of the Fourier spectrum, the corresponding amplitude ($\mathcal{A}(x)[\cdot,\cdot]$) and phase ($\mathcal{P}(x)[\cdot,\cdot]$) spectra can be expressed as,

\begin{equation}
    \mathcal{A}(x)[m,n] = \sqrt{\mathbf{Re}(\mathcal{F}(x)[m,n])^2 + \mathbf{Im}(\mathcal{F}(x)[m,n])^2}
\label{eq:amplitude_spec}
\end{equation}
\begin{equation}
    \mathcal{P}(x)[m,n] = \texttt{arctan}\left(\frac{\mathbf{Im}(\mathcal{F}(x)[m,n])}{\mathbf{Re}(\mathcal{F}(x)[m,n])}\right)
\label{eq:phase_spec}
\end{equation}

Without loss of generality, we will assume for the rest of this section that the amplitude and phase spectra are zero-centered, \textit{i.e.}, the low-frequency components (low $[m,n]$) have been shifted to the center (
lowest
frequency component is at the center).
The Fourier transform and its inverse can be calculated efficiently using the Fast Fourier Transform (FFT)~\cite{nussbaumer1981fast} algorithm. For an RGB image, we can obtain the Fourier spectrum (and $\mathcal{A}(x)[\cdot,\cdot]$ and $\mathcal{P}(x)[\cdot,\cdot]$) 
independently for each channel. For the following sections, 
although we illustrate \pst using a single-channel image,
it
can be easily extended to multi-channel (RGB) images by treating each channel independently.

\begin{figure}[t]
\centering
\includegraphics[width=\columnwidth]{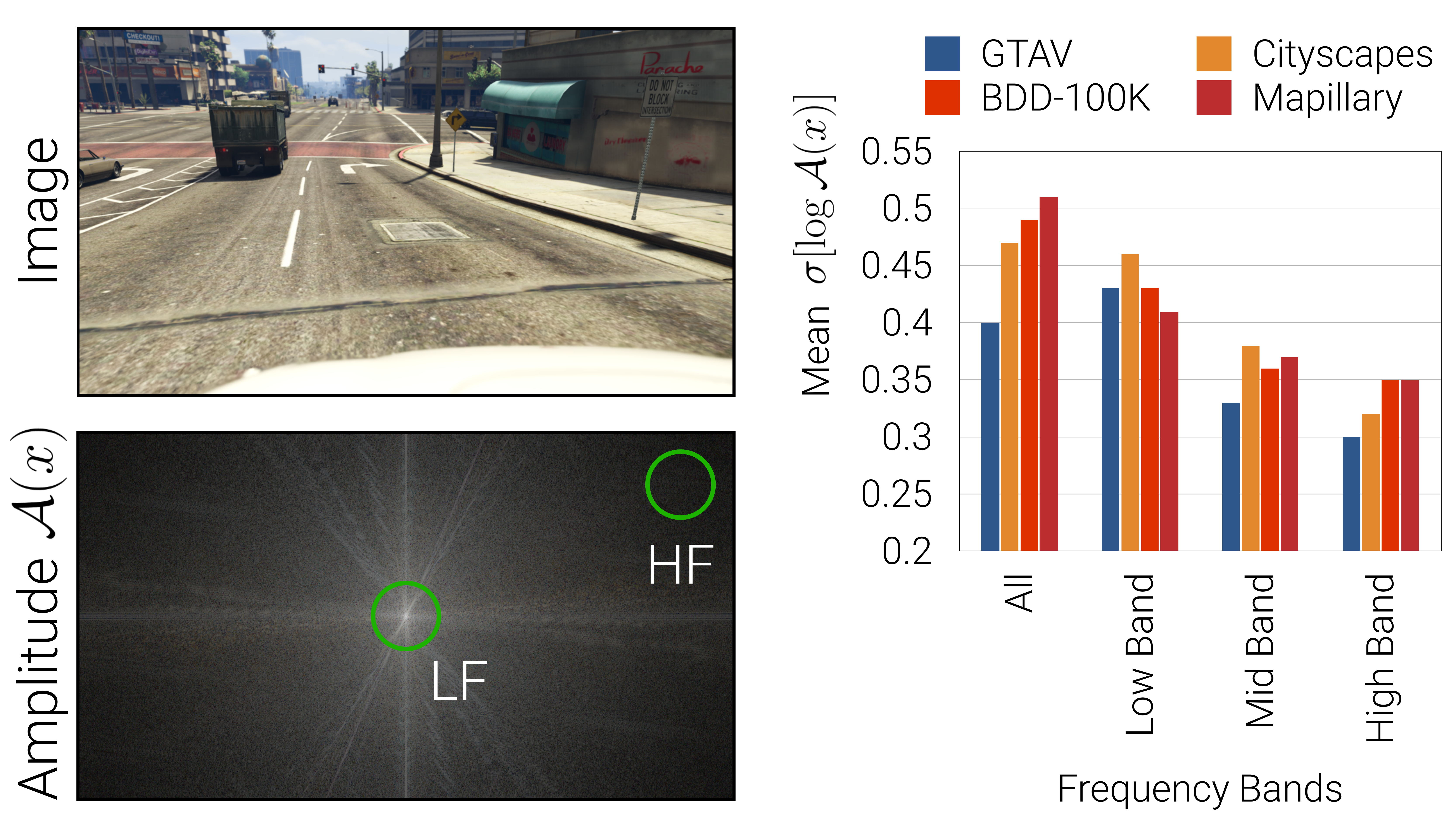}
\caption{\small\textbf{Amplitude spectrum characteristics.}  \textbf{[Left]} Sample amplitude spectrum 
(LF = Low Frequency, HF = High Frequency)
for a single channel of a synthetic image from GTAV~\cite{richter2016playing}.
Note that the amplitude spectrums tend to follow a specific pattern -- (for natural images) amplitude values tend to follow an inverse power law w.r.t. the spatial frequency~\cite{burton1987color,tolhurst1992amplitude}, \textit{i.e.},  roughly, the amplitude at frequency $f$, $\mathcal{A}(f) \propto \frac{1}{f^\gamma}$, for some $\gamma$ determined empirically.
\textbf{[Right]} Variations in amplitude values across images. Synthetic images have less variance in high-frequency components of the amplitude spectra compared to real images. }
\label{fig:amp_spectrums}
\vspace{-10pt}
\end{figure}

\subsection{Amplitude Spectrum Characteristics}

\par \noindent
\textbf{Prior Observations.} We first note that for natural images, the amplitude spectra $\mathcal{A}(x)$
has a specific structure
-- amplitude values tend to follow an inverse power law w.r.t. the spatial frequency~\cite{burton1987color,tolhurst1992amplitude}, \textit{i.e.},  roughly, the amplitude at frequency $f$, $\mathcal{A}(f) \propto \frac{1}{f^\gamma}$, for some $\gamma$ determined empirically (see Fig.~\ref{fig:amp_spectrums} [Left]). 
Moreover, as noted earlier, a considerable body of prior work~\cite{1170798,1456290,piotrowski1982demonstration,hansen2007structural} has 
shown
that the phase component of the Fourier spectrum tends to preserve the semantics of the input image,\footnote{More accurately, small variations 
in
the phase component can significantly alter the semantics of the image.} 
and the low-frequency (LF) components 
of the amplitude spectra tend to capture low-level photometric properties (illumination, etc.)~\cite{yang2020fda}. 
Based on these observations, several methods~\cite{yang2020fda,huang2021fsdr,xu2021fourier} generate augmented views by modifying only the amplitude spectra of input images, leaving the phase information ``unchanged'' -- by either copying the amplitude spectra from an image from another domain~\cite{yang2020fda} or by introducing naive uniform perturbations~\cite{xu2021fourier}.
\pst introduces a 
fine-grained
perturbation scheme for the amplitude spectra 
based on an
additional
empirical observation 
when comparing 
synthetic and real images.

\begin{figure*}[t]
\centering
\includegraphics[width=0.95\textwidth]{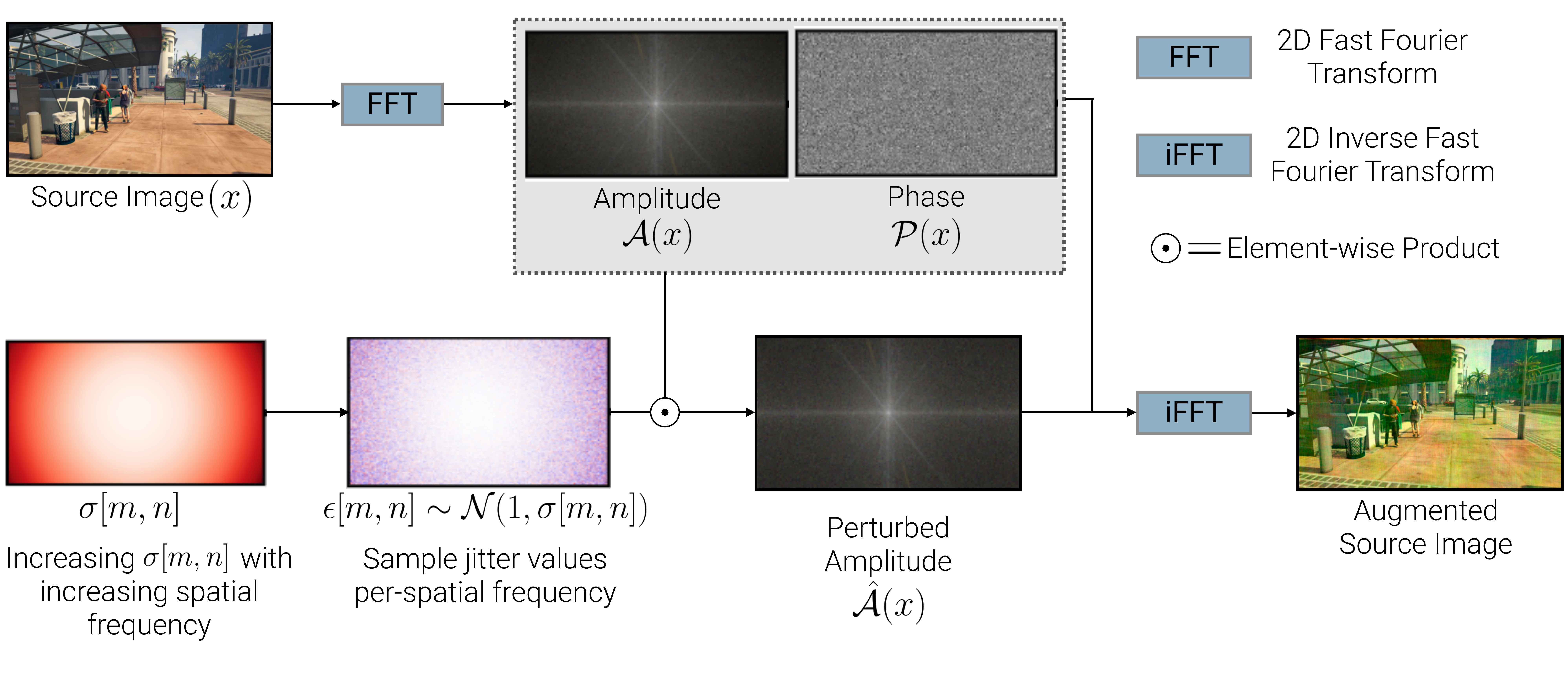}
\vspace{-15pt}
\caption{\small\textbf{\pst.} The figure outlines the augmentation pipeline involved in \pst. Given an image, we first apply a 2D Fast Fourier Transform (FFT) to obtain the amplitude and phase 
spectra.
Following this, the amplitude spectrum is perturbed as outlined in Eqns.~\ref{eq:perturb_1},~\ref{eq:fdaj_dist_settings_v1} and~\ref{eq:fdaj_dist_settings}. Finally, we use the perturbed amplitude 
and the pristine phase spectrum to recover the augmented image by 
using
inverse 2D FFT.}
\label{fig:pasta}
\vspace{-10pt}
\end{figure*}

\par \noindent
\textbf{Our Observation.} Across a set of synthetic source datasets, we make the important observation that synthetic images tend to have smaller variations in the high-frequency (HF) components of their amplitude spectrum than real images.\footnote{In Fig.~\ref{fig:amp_spectrums} [Right], for every image, upon obtaining the amplitude spectrum, we first take an element-wise logarithm. Then, for a particular frequency band (pre-defined), we compute the standard deviation of amplitude values within that band \ks{(across all the channels)}. Finally, we average these standard deviations across images to report the same in the bar plots.} 
Fig.~\ref{fig:amp_spectrums} [Right] shows the standard deviation of amplitude values for different frequency bands per-dataset with the three real datasets (for high-band of Cityscapes~\cite{cordts2016cityscapes}, BDD-100K~\cite{yu2020bdd100k}, Mapillary~\cite{neuhold2017mapillary}) being significantly larger than the synthetic dataset (high band GTAV~\cite{sankaranarayanan2018GTA}).
In appendix, we show how this phenomenon is consistent across (1) several syn-to-real shifts and (2) fine-grained frequency band discretizations. This phenomenon is likely a consequence of how synthetic images are rendered. For instance, in VisDA-C~\cite{peng2017visda}, the synthetic images are viewpoint images of 3D object models (under different lighting conditions), so it is unlikely for them to be diverse in high-frequency details. For images from GTAV~\cite{richter2016playing}, synthetic renderings can lead to contributing factors such as low texture variations – for instance, “roads” (one of the head classes in semantic segmentation) in synthetic images likely
have less high-frequency variation compared to real roads.\footnote{When \pst is applied, we find that performance on ``road'' increases by a significant margin (per-class generalization results in appendix).} Consequentially, to generalize well to real data, we would like to ensure that our augmentation strategy exposes the model to more variations in the high-frequency components of the amplitude spectrum during training. Exposing a model to such 
variations during training
allows it to be invariant to this ``nuisance" characteristic
-- which prevents overfitting to a specific syn-to-real shift.\looseness=-1

\subsection{\textbf{P}roportional \textbf{A}mplitude \textbf{S}pectrum \textbf{T}raining \textbf{A}ugmentation (\textbf{\pst})}
\label{sec:pasta}

Following the observation that synthetic data has lower amplitude variation than real images, and that the variation difference increases with larger frequencies, we introduce a new augmentation scheme, \pst, that injects variation into the amplitude spectra of synthetic images to help close the syn-to-real gap. \pst perturbs the amplitude spectra of images in a manner that is proportional to the spatial frequencies, \textit{i.e.}, higher frequencies are perturbed more than lower frequencies. If $g_{\Lambda}(\cdot)$ denotes a perturbation function that returns a perturbed amplitude $\hat{\mathcal{A}}(x)$, i.e., $\hat{\mathcal{A}}(x) = g_\Lambda(\mathcal{A}(x))$, then $g_{\Lambda}(\cdot)$ for \pst can be expressed as,

\begin{equation}
    g_{\Lambda}(\mathcal{A}(x))[m,n] = \epsilon[m,n] \mathcal{A}(x)[m,n]
    \label{eq:perturb_1}
\end{equation}
\begin{equation}
    \mbox{where }\epsilon[m,n]\sim\mathcal{N}(1,\sigma^2[m,n])
    \label{eq:fdaj_dist_settings_v1}
\end{equation}

\noindent
For every spatial frequency $[m,n]$, $\epsilon[m,n]$ ensures a ``multiplicative'' jitter interaction and is drawn from a gaussian dependent on the spatial frequency. $\sigma[m,n]$ controls the strength of perturbation applied for every 
spatial frequency.
Note that a \textit{naive} uniform perturbation to the amplitude spectrum can be applied with a constant function, $\sigma[m,n] = \beta$ for all $[m,n]$. This results in equal perturbation of all spatial frequencies. To ensure that we perturb HF components more relative to the LF ones, we need to make 
the variance ($\sigma^2[m,n]$) depend
on 
frequency ($[m,n]$). 

\input{23_tables/pasta_algo.tex}

For a given frequency we could consider a linear dependence function such as $\sigma[m,n] = 2\alpha \sqrt{\frac{m^2 + n^2}{H^2 + W^2}} + \beta$, where $2\sqrt{\frac{m^2 + n^2}{H^2 + W^2}}$ computes the normalized spatial frequency. However, in our empirical observations we found that a linear dependence on frequency does not allow for significant enough growth of perturbation as frequencies increase. Instead we propose the following polynomial function of frequency to allow for sufficient perturbation increases for the high frequency components. 
\useshortskip
\begin{equation}
    \sigma[m,n] = \underbrace{\left(2\alpha\sqrt{\frac{m^2 + n^2}{H^2 + W^2}}\right)^k}_{\text{proportional to }[m,n]} + \underbrace{\beta}_{\text{uniform}} 
    \label{eq:fdaj_dist_settings}
\end{equation}
\noindent
$\Lambda=\{\alpha, k, \beta\}$ are controllable hyper-parameters. 
Overall, $\beta$ ensures a baseline level of jitter applied to all frequencies and $\alpha,k$ govern how the perturbations grow with increasing frequencies.
Note that setting either $\alpha=0$ or $k=0$ (removing the frequency dependence) 
results in a setting where the $\sigma[m,n]$ is the same across all $[m,n]$. In appendix, we 
verify that
\pst augmentation
increases the variance metric measured in Fig.~\ref{fig:amp_spectrums} [Right] for synthetic images across fine-grained frequency band discretizations.

The steps involved in obtaining a \pst augmented view are summarized in Alg.~\ref{algo:pasta} and Fig.~\ref{fig:pasta}. Given an input image, we first obtain the fourier, amplitude and phase spectra via 2D FFT
and then
perturb the amplitude spectrum (while ensuring 
stronger perturbations for HF components)
according to Eqns.~\ref{eq:perturb_1},~\ref{eq:fdaj_dist_settings_v1} and~\ref{eq:fdaj_dist_settings}. Finally, given the perturbed amplitude spectrum and the pristine phase spectrum, we retrieve the augmented image via inverse 2D FFT. In the next section we empricially validate our augmentation strategy.

%% file: 23_tables/pasta_algo.tex
\begin{figure}[t]
\begin{algorithm}[H]
\caption{\pst Augmented Views}
\label{algo:pasta}
\begin{algorithmic}[1]
\State \textbf{Input:} $x\in\mathbb{R}^{H\times W}$ \Comment{\textcolor{blue}{Synthetic image (single channel)}}
\State \textbf{\pst parameters:} $\alpha,k,\beta$ \Comment{\textcolor{blue}{Set values}}
\State $\mathcal{F}(x)$ $\gets$ FFT $(x)$ \Comment{\textcolor{blue}{Obtain Fourier spectrum}}
\State $\mathcal{A}(x)$ $\gets$ Abs $(\mathcal{F}(x))$ \Comment{\textcolor{blue}{Obtain amplitude spectrum}}
\State $\mathcal{P}(x)$ $\gets$ Ang $(\mathcal{F}(x))$ \Comment{\textcolor{blue}{Obtain phase spectrum}}
\State $\mathcal{A}(x)$ $\gets$ FFTShift $(\mathcal{A}(x))$ \Comment{\textcolor{blue}{Zero-center amplitude}}
\State $\sigma_{H\times W}$ $\gets$ Meshgrid$([-H/2,H/2], [-W/2,W/2])$
\For{$m\in[-H/2,H/2]$} \Comment{\textcolor{brown}{Vectorized in practice}}
    \For{$n\in[-W/2,W/2]$}
        \State $\sigma[m,n] \gets \left(2\alpha\sqrt{\frac{m^2 + n^2}{H^2 + W^2}}\right)^k + \beta$  
    \EndFor
\EndFor
\State $\epsilon_{H\times W}$ $\sim\mathcal{N}(1,\sigma^2_{H\times W})$ \Comment{\textcolor{blue}{Draw perturbations}}
\State $\hat{\mathcal{A}}(x) \gets \epsilon \odot \mathcal{A}(x)$ \Comment{\textcolor{blue}{Multiplicative jitter}}
\State $\hat{\mathcal{A}}(x)$ $\gets$ FFTShift $(\hat{\mathcal{A}}(x))$
\State $\hat{x} \gets$ Inverse-FFT $(\hat{\mathcal{A}}(x), \mathcal{P}(x))$ \Comment{\textcolor{blue}{Augmented Image}}
\end{algorithmic}
\end{algorithm}
\vspace{-5pt}
\end{figure}

%% file: sections/experiments_new.tex
We conduct synthetic-to-real generalization experiments across three tasks -- Semantic Segmentation (SemSeg), Object Detection (ObjDet) and Object Recognition (ObjRec). \looseness=-1
\subsection{Datasets and Shifts}
\par \noindent
\textbf{Semantic Segmentation.} For SemSeg, we use GTAV~\cite{richter2016playing} as our synthetic source dataset with $\sim25$k ground-view images and $19$ annotated classes, which are compatible with the classes in real target datasets -- Cityscapes~\cite{cordts2016cityscapes}, BDD100K~\cite{yu2020bdd100k} and Mapillary~\cite{neuhold2017mapillary}. We train our models on the training split of the synthetic sources, evaluate on the validation splits of the real targets and report performance using mIoU (mean intersection over union). We train SegFormer and HRDA (source-only) on the entirety of GTAV.

\par \noindent
\textbf{Object Detection.} For ObjDet, we use Sim10K~\cite{johnson2016driving} as our synthetic source dataset with $\sim10$k street-view images (from GTAV~\cite{richter2016playing}) and Cityscapes~\cite{cordts2016cityscapes} as our (real) target dataset. Following prior work~\cite{khindkar2022miss}, we train on the entirety of Sim10K to detect instances of ``car'' and report performance on the val split of Cityscapes using mAP@50 (mean average precision at an IoU threshold of 0.5).

\par \noindent
\textbf{Object Recognition.} For ObjRec, we use the VisDA-2017~\cite{peng2017visda} (syn$\rightarrow$real) image-classification benchmark with $\sim152$k synthetic images (3D renderings of objects) and $\sim55$k real images (from COCO~\cite{lin2014microsoft}) across $12$ classes. We use (class-balanced) accuracy as our evaluation metric. 

\subsection{Models and Baselines}
\label{sec:implementation_details}
\par \noindent
\textbf{Models.} We use DeepLabv3+~\cite{chen2018encoder} (with backbones ResNet-50~\cite{he2016deep}, ResNet-101~\cite{he2016deep}), SegFormer~\cite{xie2021segformer}
and 
HRDA~\cite{hoyer2022hrda} (both with MiT-B5 backbones) architectures for SemSeg experiments. For ObjDet, we use the Faster-RCNN~\cite{ren2015faster} architecture with ResNet-50 and ResNet-101 backbones. For ObjRec, we use ViT-B/16~\cite{dosovitskiy2021an} and ResNet-101 architectures, with both supervised and self-supervised (DINO~\cite{caron2021emerging}) initializations. We set
\pst hyper-parameters $(\alpha=3.0, k = 2.0, \beta=0.25)$ for SemSeg, ObjDet and ObjRec across shifts, backbones
and apply it in conjunction with consistent geometric and photometric augmentations per task. We provide more details in appendix.

\par \noindent
\textbf{Points of Comparison.} In addition to prior work in syn-to-real generalization, we also compare \pst with other augmentation strategies -- (1) RandAugment (RandAug)~\cite{cubuk2020randaugment} and (2) Photometric Distortion (PD)~\cite{chen2019mmdetection}. The sequence of operations in PD to augment input images are – randomized brightness, randomized contrast, RGB$\rightarrow$HSV
conversion, randomized saturation \& hue changes, HSV$\rightarrow$RGB conversion, randomized contrast, and randomized channel swap.

%% file: sections/results_new.tex
\input{23_tables/semseg_baseline_improvement.tex}
\input{23_tables/detection_baseline_improvement.tex}
\input{23_tables/classification_baseline_improvement.tex}
\subsection{Synthetic-to-Real Generalization Results}
\label{sec:main_pasta_results}
Our syn-to-real generalization results for Semantic Segmentation (SemSeg), Object Detection (ObjDet) and Object Recognition (ObjRec) are summarized in Tables.~\ref{table:semseg_baseline_improvements},~\ref{table:detection_baseline_improvements},~\ref{table:r101_visda_cls_res_baseline}
~\ref{table:semseg_sota},~\ref{table:detection_sota},~\ref{table:semseg_complementary},
and~\ref{table:semseg_cnn_transformer}. We discuss these results below.
\par \noindent
\textbf{$\triangleright$ \pst considerably improves a vanilla baseline.} Tables.~\ref{table:semseg_baseline_improvements} and~\ref{table:detection_baseline_improvements} show the improvements offered by \pst for Semantic Segmentation (SemSeg) and Object Detection (ObjDet) respectively when applied to a vanilla baseline. For SemSeg (see Table.~\ref{table:semseg_baseline_improvements}), we find that \pst improves a baseline DeepLabv3+ model by $13+$ absolute mIoU points (see rows 1, 3, 4 and 5) across R-50 and R-101 backbones. Furthermore, these improvements are obtained consistently across \textit{all} real target datasets. Similarly, for ObjDet (Table.~\ref{table:detection_baseline_improvements}), \pst offers absolute improvements of $11+$ mAP points for a Faster-RCNN baseline across R-50 and R-101 backbones (see rows 1, 4, 7 and 10). We further note that for SemSeg, \pst outperforms RandAugment~\cite{cubuk2020randaugment}, a competing augmentation strategy. For ObjDet, \pst either outperforms (R-50; rows 3, 4) 
or is competitive with RandAug (R-101; rows 9, 10).
For ObjRec (see Table.~\ref{table:r101_visda_cls_res_baseline}), we find that \pst significantly improves a vanilla baseline across multiple architectures -- for R-101 (rows 1, 2) and ViT-B/16 (rows 3, 4) -- and initializations -- for supervised (rows 3, 4) and DINO~\cite{caron2021emerging} (rows 5, 6) ViT-B/16 initializations.

\input{23_tables/semseg_pasta_sota.tex}
\input{23_tables/detection_sota.tex}

\par \noindent
\textbf{$\triangleright$ \pst outperforms state-of-the-art generalization methods.} Table.~\ref{table:semseg_sota} shows how applying \pst to a baseline outperforms existing generalization methods for SemSeg. For instance, Baseline + \pst outperforms IBN-Net~\cite{pan2018two}, ISW~\cite{choi2021robustnet}, DRPC~\cite{yue2019domain}, ASG~\cite{chen2020automated}, CSG~\cite{chen2021contrastive}, WEDGE~\cite{kim2023wedge}, FSDR~\cite{huang2021fsdr}, WildNet~\cite{lee2022wildnet}, DIRL~\cite{xu2022dirl} \& SHADE~\cite{zhao2022shade} in terms of average mIoU across real targets (for both R-50 and R-101).
We would like to note that DRPC, ASG, CSG, WEDGE, FSDR, WildNet \& SHADE (for R-101) use either more synthetic or real data (the entirety of GTAV~\cite{richter2016playing} or additional datasets) or different base architectures, making these comparisons unfair \textit{to} \pst.
Overall, when compared to prior work, Baseline + \pst, achieves state-of-the-art results across both backbones without the use of any specialized components, task-specific design, or changes to learning rules.
For ObjDet, in Table.~\ref{table:detection_sota}, we find that combining RandAug + \pst sets new state-of-the-art on 
Sim10K~\cite{johnson2016driving}$\rightarrow$Cityscapes~\cite{cordts2016cityscapes}. 

\par \noindent
\textbf{$\triangleright$ \pst outperforms state-of-the-art adaptation methods.} In Table.~\ref{table:detection_sota}, we find that both Baseline + \pst and Baseline + RandAug + \pst significantly outperform state-of-the-art adaptive object detection method, AWADA~\cite{menke2022awada} (rows 2, 3 and 7). Note that unlike the methods in rows 4-7, \pst does not have access to real images during training!

\input{23_tables/semseg_pasta_complementary.tex}

\par \noindent
\textbf{$\triangleright$ \pst is complementary to existing generalization methods.} In addition to ensuring that a baseline model improves over existing methods, we find that
\pst is also complementary to existing generalization methods. For SemSeg, in Table.~\ref{table:semseg_complementary}, we find that applying \pst significantly improves generalization performance ($5+$ absolute mIoU points) of IBN-Net~\cite{pan2018two} and ISW~\cite{choi2021robustnet} across R-50 and R-101. For ObjDet, we find that \pst is complementary to existing augmentation methods (PD~\cite{chen2019mmdetection} and RandAug~\cite{cubuk2020randaugment}; rows 5, 6, 11 and 12 in Table.~\ref{table:detection_baseline_improvements}), with RandAug + \pst setting state-of-the-art on Sim10K~\cite{johnson2016driving}$\rightarrow$Cityscapes~\cite{cordts2016cityscapes}. 
In appendix, we applied \pst to CSG~\cite{chen2021contrastive}, a state-of-the-art generalization method for ObjRec on VisDA-C~\cite{peng2017visda}
CSG already utilizes RandAugment, so we tested \pst in two settings: with and without RandAugment. In both scenarios, incorporating \pst led to improved performance.

\subsection{Analyzing \pst}
\label{sec:main_pasta_analysis}

\par
\begin{figure}[t]
\centering
\includegraphics[width=\linewidth]{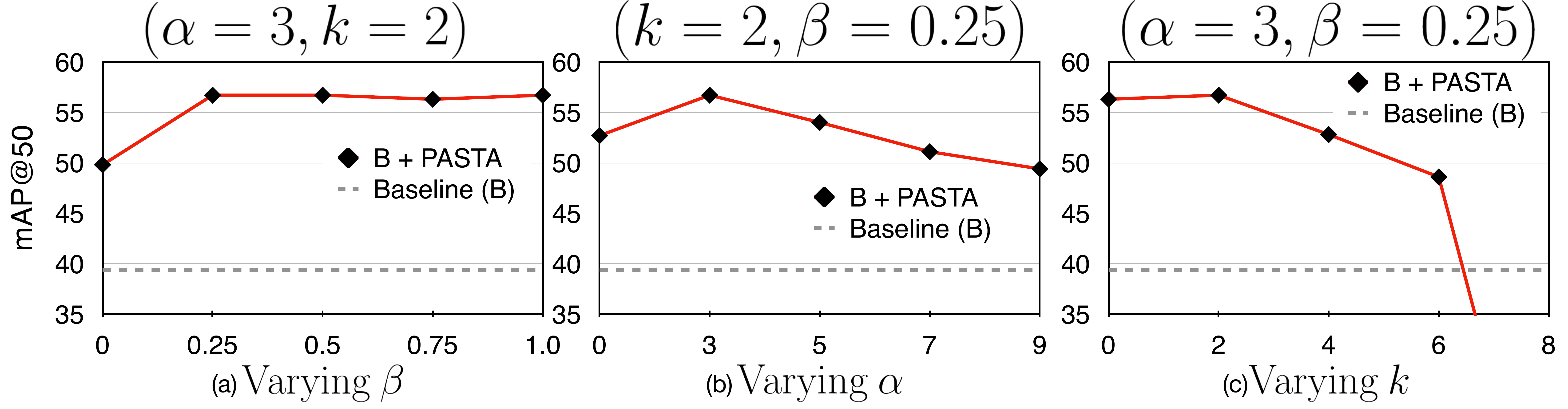}
\caption{\small\textbf{Sensitivity of $\alpha$, $\beta$, $k$ in \pst for Object Detection.} Faster-RCNN (ResNet-50) models trained on Sim10K (S) and evaluated on Cityscapes (C). We vary $\alpha,k$ and $\beta$ for \pst within the sets $\{0, 3, 5, 7, 9\}$, $\{0, 2, 4, 6, 8\}$ and $\{0, 0.25, 0.5, 0.75, 1.0\}$.}
\label{fig:pasta_ablations}
\vspace{-15pt}
\end{figure}

 \noindent
\textbf{$\triangleright$ Sensitivity of \pst to $\alpha$, $k$ and $\beta$.} While $\beta$ provides a baseline level of uniform jitter in the frequency domain, $\alpha,k$ govern the degree of monotonicity applied to the perturbations. To assess the sensitivity of \pst to $\alpha,k,\beta$, in Fig.~\ref{fig:pasta_ablations}, we conduct experiments for ObjDet where we vary one hyper-parameter while freezing the other two. 
We find that performance is stable when $\beta$ exceeds a certain threshold.
For $\alpha$, we find that while performance drops with increasing $\alpha$, worst generalization performance is still significantly above baseline. More importantly, we find generalization improvements offered by \pst are sensitive to ``extreme'' values of $k$. Qualitatively, overly high values of $\alpha,k$ lead to augmented views which have their semantic content significantly occluded, thereby resulting in poor generalization. As a rule of thumb, for a vanilla baseline that is not designed specifically for syn-to-real generalization, we find that restricting 
$k\in[1,4]$ leads to stable improvements.

\par \noindent
\textbf{$\triangleright$ \pst vs other frequency-based augmentation strategies.} Prior work has also considered augmenting images in the Fourier domain for syn-to-real generalization (FSDR~\cite{huang2021fsdr}), multi-source domain generalization (FACT~\cite{xu2021fourier}), domain adaptation (FDA~\cite{yang2020fda}) and robustness (APR~\cite{chen2021amplitude}). 
Row 17 in Table.~\ref{table:semseg_sota} already shows how \pst outperforms FSDR. In appendix, 
we show
that for baseline DeepLabv3+ (R-50) SemSeg models on GTAV$\rightarrow$Real 
\pst outperforms 
FACT,
FDA and APR.

\par \noindent
\textbf{$\triangleright$ Does monotonicity matter in \pst?} As stated earlier, a key insight while designing \pst was to perturb the high-frequency components in the amplitude spectrum more relative to the low-frequency ones. This monotonicity is governed by the choice of $\alpha,k$ in Eqn.~\ref{eq:fdaj_dist_settings} ($\beta$ applies a uniform level of jitter to all frequency components). To assess the importance of this monotonic setup, we compare generalization improvements offered by \pst by comparing $\alpha=0$ (uniform) and $\alpha=3$ (monotonically increasing) settings. For SemSeg, for a baseline DeepLabv3+ (R-50) model, we find that $\alpha=3$ improves over $\alpha=0$ by $3.89$ absolute 
mIoU points (see Table.~\ref{table:semseg_baseline_improvements} for experimental setting). Similarly, for ObjDet, for a baseline Faster-RCNN (R-50) model, we find that $\alpha=3$ improves of $\alpha=0$ by $3.4$ absolute 
mAP points (see Table.~\ref{table:detection_baseline_improvements} for experimental setting). Additionally, reversing the monotonic trend 
(LF perturbed more than HF)
leads to significantly worse generalization performance for SemSeg -- $8.90$ Avg. Real mIoU, dropping even below vanilla baseline performance of $27.42$ mIoU.

\input{23_tables/semseg_cnn_transformer.tex}

\par \noindent
\textbf{$\triangleright$ Does \pst help for transformer based architectures?} Our key syn-to-real generalization results across SemSeg
and ObjDet
are 
with CNN backbones. Following the recent surge of interest in introducing transformers to vision tasks~\cite{dosovitskiy2021an}, we also conduct experiments to assess if syn-to-real generalization improvements offered by \pst generalize beyond CNNs.
In Table.~\ref{table:r101_visda_cls_res_baseline}, we show how applying \pst improves syn-to-real generalization performance of ViT-B/16 baselines for supervised (rows 3, 4) and self-supervised DINO~\cite{caron2021emerging} initializations (rows 5, 6).
In Table.~\ref{table:semseg_cnn_transformer}, we consider SegFormer~\cite{xie2021segformer} and HRDA~\cite{hoyer2022hrda} (source-only), 
recent transformer based segmentation frameworks,
and check syn-to-real generalization performance when trained on GTAV~\cite{richter2016playing} 
and evaluated on Cityscapes~\cite{cordts2016cityscapes}. We find that applying \pst improves performance of a vanilla baseline 
($6+$ and $4+$ mIoU points for SegFormer and HRDA respectively).

To summarize, 
from our experiments, we 
find
that \pst serves as a simple and effective \textit{plug-and-play} augmentation strategy 
for
training on synthetic data that
\begin{packed_itemize}
    \item[--] \textbf{provides strong out-of-the-box generalization performance} -- enables a baseline method to outperform existing generalization approaches
    \item[--] \textbf{is complementary to existing generalization methods} -- applying \pst to existing methods or augmentation strategies leads to improvements
    \item[--] \textbf{is applicable across tasks, backbones and shifts} -- \pst leads to improvements across SemSeg, ObjDet, ObjRec for multiple backbones for 
    five syn-to-real shifts
\end{packed_itemize}

%% file: 23_tables/semseg_baseline_improvement.tex
\begin{table}[t]
\footnotesize
\centering
\setlength{\tabcolsep}{3.0pt}
\begin{center}
\resizebox{\columnwidth}{!}{
\begin{tabular}{lccccc}
\toprule
\multirow{2}{*}{\textbf{Method}} & \multicolumn{4}{c}{\textbf{Real mIoU} $\uparrow$}\\ \cmidrule{2-6}
 & G$\rightarrow$C &G$\rightarrow$B & G$\rightarrow$M & Avg & $\Delta$\\
\midrule
ResNet-50 & & & &\\
\midrule
\multicolumn{1}{l}{\texttt{1} Baseline (B)~\cite{choi2021robustnet}$^*$} & 28.95 & 25.14 & 28.18 & 27.42 \\
\multicolumn{1}{l}{\texttt{2} B + RandAug~\cite{cubuk2020randaugment}} & 31.89 & 38.28 & 34.54 & 34.54\tiny{\std{0.57}}& {\color{blue}+7.12}\\
\multicolumn{1}{l}{\texttt{3} B + \pst} & \textbf{44.12} & \textbf{40.19} & \textbf{47.11} &  \textbf{43.81\tiny{\std{0.74}}} & {\color{blue}+16.39}\\
\midrule
ResNet-101 & & & &\\
\midrule
\multicolumn{1}{l}{\texttt{4} Baseline (B)~\cite{choi2021robustnet}$^*$} & 32.97 & 30.77 & 30.68 & 31.47 \\
\multicolumn{1}{l}{\texttt{5} B + \pst} & \textbf{45.33} & \textbf{42.32 }& \textbf{48.60} & \textbf{45.42\std{0.14}} & {\color{blue}+13.95}\\
\bottomrule
\end{tabular}}
\caption{\small\textbf{\pst considerably improves a (SemSeg) baseline.} Semantic Segmentation DeepLabv3+ models trained on GTAV (G) and evaluated on $\{$Cityscapes (C), BDD100K (B), Mapillary (M)$\}$. $^*$ indicates numbers drawn from published manuscripts. 
\textbf{Bold} indicates best. $\Delta$ indicates (absolute) improvement over Baseline. \textcolor{bluebell}{$\pm$} indicates the standard deviation across 3 random seeds.}
\label{table:semseg_baseline_improvements}
\end{center}
\vspace{-15pt}
\end{table}

%% file: 23_tables/detection_baseline_improvement.tex
\begin{table}[t]
\footnotesize
\centering
\setlength{\tabcolsep}{6.5pt}
\begin{center}
\resizebox{\columnwidth}{!}{
\begin{tabular}{lcc}
\toprule
\textbf{Method} & \textbf{mAP@50} $\uparrow$ & $\Delta$ \\
\midrule
ResNet-50 & S$\rightarrow$C\\
\midrule
\texttt{1} Baseline & 39.4 & \\
\texttt{2} Baseline + PD~\cite{chen2019mmdetection} & 51.5 & {\color{blue} +12.1}\\
\texttt{3} Baseline + RandAug~\cite{cubuk2020randaugment} & 52.8 &  {\color{blue} +13.4}\\
\texttt{4} Baseline + \pst & 56.3 &  {\color{blue} +16.9}\\
\texttt{5} Baseline + PD + \pst & 58.0 &  {\color{blue} +18.6}\\
\texttt{6} Baseline + RandAug + \pst & \textbf{58.3} &  {\color{blue} +18.9}\\
\midrule
ResNet-101 \\
\midrule
\texttt{7} Baseline & 43.3 &  \\
\texttt{8} Baseline + PD~\cite{chen2019mmdetection} & 52.2 &  {\color{blue} +8.9}\\
\texttt{9} Baseline + RandAug~\cite{cubuk2020randaugment} & 57.2 &  {\color{blue} +13.9}\\
\texttt{10} Baseline + \pst & 55.2 & {\color{blue} +11.9}\\
\texttt{11} Baseline + PD + \pst & 56.6 &  {\color{blue} +13.3}\\
\texttt{12} Baseline + RandAug + \pst & \textbf{59.9} &  {\color{blue} +16.6}\\
\bottomrule
\end{tabular}}
\caption{\small\textbf{\pst considerably improves a (ObjDet) baseline.} Object Detection Faster-RCNN models trained on Sim10K (S) and evaluated on Cityscapes (C). $^*$ indicates numbers drawn from published manuscripts. 
Highest is \textbf{Bolded}. $\Delta$ indicates (absolute) improvement over Baseline.}
\label{table:detection_baseline_improvements}
\end{center}
\vspace{-15pt}
\end{table}

%% file: 23_tables/classification_baseline_improvement.tex
\begin{table}[t]
\footnotesize
\centering
\setlength{\tabcolsep}{2.5pt}
\begin{center}
\resizebox{\columnwidth}{!}{
\begin{tabular}{lcccc}
\toprule
\multirow{2}{*}{\textbf{Method}} & \multirow{2}{*}{\textbf{Arch.}} & \multirow{2}{*}{\textbf{Init.}} & \textbf{Accuracy} $\uparrow$ &  $\Delta$\\
 & & & Syn$\rightarrow$Real \\
\midrule
\texttt{1} Baseline & ResNet-101 & Sup. & 47.22 &\\
\texttt{2} Baseline + \pst & ResNet-101 & Sup. &  \textbf{54.39} & {\color{blue} +7.17}\\
\midrule
\texttt{3} Baseline & ViT-B/16 & Sup. &  56.06 &\\
\texttt{4} Baseline + \pst & ViT-B/16 & Sup. &  \textbf{58.08} &{\color{blue} +2.02}\\
\midrule
\texttt{5} Baseline & ViT-B/16 & DINO~\cite{caron2021emerging} & 60.93 &\\
\texttt{6} Baseline + \pst & ViT-B/16 & DINO~\cite{caron2021emerging} & \textbf{63.55} &{\color{blue} +2.62}\\
\bottomrule
\end{tabular}}
\caption{\small\textbf{\pst considerably improves a (ObjRec) baseline.} Classification models trained on the synthetic source split of VisDA-C and evaluated on the real-split of VisDA-C. 
\textbf{Bold} is best. $\Delta$ indicates (absolute) improvement over Baseline. Sup.$=$ Supervised.}
\label{table:r101_visda_cls_res_baseline}
\end{center}
\vspace{-15pt}
\end{table}

%% file: 23_tables/semseg_pasta_sota.tex
\begin{table}[t]
\footnotesize
\centering
\setlength{\tabcolsep}{3.5pt}
\begin{center}
\resizebox{\columnwidth}{!}{
\begin{tabular}{lccccc}
\toprule
\multirow{2}{*}{\textbf{Method}} & \multicolumn{4}{c}{\textbf{Real mIoU} $\uparrow$}\\ \cmidrule{2-6}
 & G$\rightarrow$C &G$\rightarrow$B & G$\rightarrow$M & Avg & $\Delta$\\
\midrule
ResNet-50 & & & &\\
\midrule
\multicolumn{1}{l}{\texttt{1} IBN-Net~\cite{pan2018two}$^*$} & 33.85 & 32.30 & 37.75 & 34.63 \\
\multicolumn{1}{l}{\texttt{2} ISW~\cite{choi2021robustnet}$^*$} & 36.58 & 35.20 & 40.33 & 37.37 \\
\rowcolor{Gray}
\multicolumn{1}{l}{\texttt{3} DRPC~\cite{yue2019domain}$^*$} & 37.42 & 32.14 & 34.12 & 34.56 &\\
\rowcolor{Gray}
\multicolumn{1}{l}{\texttt{4} WEDGE~\cite{kim2023wedge}$^*$} & 38.36 & 37.00 & 44.82 & 40.06 &\\
\rowcolor{Gray}
\multicolumn{1}{l}{\texttt{5} ASG~\cite{chen2020automated}$^*$} & 31.89 & N/A & N/A & N/A &\\
\rowcolor{Gray}
\multicolumn{1}{l}{\texttt{6} CSG~\cite{chen2021contrastive}$^*$} & 35.27 & N/A & N/A & N/A &\\
\rowcolor{Gray}
\multicolumn{1}{l}{\texttt{7} WildNet~\cite{lee2022wildnet}$^*$} & 44.62 & 38.42 & 46.09 & 43.04 &\\
\multicolumn{1}{l}{\texttt{8} DIRL~\cite{xu2022dirl}$^*$} & 41.04 & 39.15 & 41.60 & 40.60 \\
\multicolumn{1}{l}{\texttt{9} SHADE~\cite{zhao2022shade}$^*$} & \textbf{44.65} & 39.28 & 43.34 & 42.42 \\
\multicolumn{1}{l}{\texttt{10} B + \pst} & 44.12 & \textbf{40.19} & \textbf{47.11} &  \textbf{43.81\tiny{\std{0.74}}} & {\color{blue}+0.77}\\
\midrule
ResNet-101 & & & &\\
\midrule
\multicolumn{1}{l}{\texttt{11} IBN-Net~\cite{pan2018two}$^*$} & 37.37 & 34.21 & 36.81 & 36.13 \\
\multicolumn{1}{l}{\texttt{12} ISW~\cite{choi2021robustnet}$^*$} & 37.20 & 33.36 & 35.57 & 35.58 \\
\rowcolor{Gray}
\multicolumn{1}{l}{\texttt{13} DRPC~\cite{yue2019domain}$^*$} & 42.53 & 38.72 & 38.05 & 39.77 &\\
\rowcolor{Gray}
\multicolumn{1}{l}{\texttt{14} WEDGE~\cite{kim2023wedge}$^*$} & 45.18 & 41.06&  48.06& 44.77  &\\
\rowcolor{Gray}
\multicolumn{1}{l}{\texttt{15} ASG~\cite{chen2020automated}$^*$} &  32.79 & N/A & N/A & N/A &\\
\rowcolor{Gray}
\multicolumn{1}{l}{\texttt{16} CSG~\cite{chen2021contrastive}$^*$} & 38.88 & N/A & N/A & N/A &\\
\rowcolor{Gray}
\multicolumn{1}{l}{\texttt{17} FSDR~\cite{huang2021fsdr}$^*$} & 44.80 & 41.20 & 43.40 & 	43.13 &\\
\rowcolor{Gray}
\multicolumn{1}{l}{\texttt{18} WildNet~\cite{lee2022wildnet}$^*$} & 45.79 & 41.73 & 47.08 & 44.87 &\\
\rowcolor{Gray}
\multicolumn{1}{l}{\texttt{19} SHADE~\cite{zhao2022shade}$^*$} & \textbf{46.66} & \textbf{43.66} & 45.50 & 45.27 &\\
\multicolumn{1}{l}{\texttt{20} B + \pst} & 45.33 & 42.32& \textbf{48.60} & \textbf{45.42\std{0.14}} & {\color{blue}+0.15}\\
\bottomrule
\end{tabular}}
\caption{\small\textbf{\pst outperforms SOTA (SemSeg) generalization methods.} Semantic Segmentation DeepLabv3+ models trained on GTAV (G) and evaluated on $\{$Cityscapes (C), BDD100K (B), Mapillary (M)$\}$. $^*$ indicates numbers drawn from published manuscripts. 
\textbf{Bold} is best. $\Delta$ indicates (absolute) improvement over SOTA. \textcolor{bluebell}{$\pm$} indicates the standard deviation across 3 random seeds. Rows highlighted in \unfair{gray} use different base architectures and / or extra 
training data 
and are included for completeness. 
B $=$ Baseline.}
\label{table:semseg_sota}
\end{center}
\vspace{-15pt}
\end{table}

%% file: 23_tables/detection_sota.tex
\begin{table}[t]
\footnotesize
\centering
\setlength{\tabcolsep}{4.5pt}
\begin{center}
\resizebox{\columnwidth}{!}{
\begin{tabular}{lccc}
\toprule
\textbf{Method} & Real Data & \textbf{mAP@50} $\uparrow$ & $\Delta$ \\
\midrule
Generalization & & S$\rightarrow$C\\
\midrule
\texttt{1} Baseline (B) & \xmark & 43.3 &  \\
\texttt{2} B + \pst & \xmark & 55.2 &  {\color{blue} +11.9}\\
\texttt{3} B + RandAug + \pst & \xmark & \textbf{59.9} &  {\color{blue} +16.6}\\
\midrule
(Unsupervised) Adaptation \\
\midrule
\rowcolor{Gray}
\multicolumn{1}{l}{\texttt{4} EPM~\cite{hsu2020every}$^*$} & \cmark & 51.2 & {\color{blue} +7.9}\\
\rowcolor{Gray}
\multicolumn{1}{l}{\texttt{5} Faster-RCNN w/ rot~\cite{wang2021robustobjdet}$^*$} & \cmark &  52.4 & {\color{blue} +9.1}\\
\rowcolor{Gray}
\multicolumn{1}{l}{\texttt{6} ILLUME~\cite{khindkar2022miss}$^*$} & \cmark &  53.1 & {\color{blue} +9.8}\\
\rowcolor{Gray}
\multicolumn{1}{l}{\texttt{7} AWADA~\cite{menke2022awada}$^*$} & \cmark &  53.2 & {\color{blue} +9.9}\\
\cmidrule{1-4}
\rowcolor{red!15}
\multicolumn{1}{l}{\texttt{8} Faster-RCNN (Oracle)~\cite{wang2021robustobjdet}$^*$} & \cmark & 70.4 & {\color{blue} +27.1}\\
\bottomrule
\end{tabular}}
\caption{\small\textbf{\pst outperforms SOTA (ObjDet) adaptation methods.} Object Detection Faster-RCNN (ResNet-101) models trained on Sim10K (S) and evaluated on Cityscapes (C). $^*$ indicates numbers drawn from published manuscripts. 
\textbf{Bold} is best.
Rows in \unfair{gray} are adaptation methods with access to unlabeled real data. Rows in \md{red} have access to labeled real data.
$\Delta$ indicates (absolute) improvement over Baseline (B). \pst leads to outperforming adaptation methods without any access to real data during training.
}
\label{table:detection_sota}
\end{center}
\vspace{-10pt}
\end{table}

%% file: 23_tables/semseg_pasta_complementary.tex
\begin{table}[t]
\footnotesize
\centering
\setlength{\tabcolsep}{3.5pt}
\begin{center}
\resizebox{\columnwidth}{!}{
\begin{tabular}{lccccc}
\toprule
\multirow{2}{*}{\textbf{Method}} & \multicolumn{4}{c}{\textbf{Real mIoU} $\uparrow$}\\ \cmidrule{2-6}
 & G$\rightarrow$C &G$\rightarrow$B & G$\rightarrow$M & Avg & $\Delta$\\
\midrule
ResNet-50 & & & &\\
\midrule
\multicolumn{1}{l}{\texttt{1} IBN-Net~\cite{pan2018two}$^*$} & 33.85 & 32.30 & 37.75 & 34.63 \\
\multicolumn{1}{l}{\texttt{2} IBN-Net + \pst} & \textbf{41.90} & \textbf{41.46} & \textbf{45.88} & \textbf{43.08\tiny{\std{0.37}}}& {\color{blue} +8.45}\\
\multicolumn{1}{l}{\texttt{3} ISW~\cite{choi2021robustnet}$^*$} & 36.58 & 35.20 & 40.33 & 37.37 \\
\multicolumn{1}{l}{\texttt{4} ISW + \pst} & \textbf{42.13} & \textbf{40.95} & \textbf{45.67} & \textbf{42.91\tiny{\std{0.27}}}& {\color{blue} +5.54}\\
\midrule
ResNet-101 & & & &\\
\midrule
\multicolumn{1}{l}{\texttt{5} IBN-Net~\cite{pan2018two}$^*$} & 37.37 & 34.21 & 36.81 & 36.13 \\
\multicolumn{1}{l}{\texttt{6} IBN-Net + \pst} & \textbf{43.64} & \textbf{42.46} & \textbf{47.51} & \textbf{44.54\std{0.89}} & {\color{blue} +8.41}\\
\multicolumn{1}{l}{\texttt{7} ISW~\cite{choi2021robustnet}$^*$} & 37.20 & 33.36 & 35.57 & 35.58 \\
\multicolumn{1}{l}{\texttt{8} ISW + \pst} & \textbf{44.46} & \textbf{43.02} & \textbf{47.35} & \textbf{44.95\std{0.21}} & {\color{blue} +9.37}\\
\bottomrule
\end{tabular}}
\vspace{-2pt}
\caption{\small\textbf{\pst is complementary to existing (SemSeg) generalization methods.} Semantic Segmentation generalization methods (DeepLabv3+ models) trained on GTAV (G) and evaluated on $\{$Cityscapes (C), BDD100K (B), Mapillary (M)$\}$. $^*$ indicates numbers drawn from published manuscripts. 
\textbf{Bold} is best. $\Delta$ indicates (absolute) improvement over the base generalization method. 
\textcolor{bluebell}{$\pm$} indicates the standard deviation across 3 random seeds.
}
\label{table:semseg_complementary}
\end{center}
\vspace{-20pt}
\end{table}

%% file: 23_tables/semseg_cnn_transformer.tex
\begin{table}[t]
\footnotesize
\centering
\setlength{\tabcolsep}{6.5pt}
\begin{center}
\resizebox{\columnwidth}{!}{
\begin{tabular}{lcc}
\toprule
\textbf{Method} & \textbf{G$\rightarrow$C mIoU} $\uparrow$ & $\Delta$\\
\midrule
DeepLabv3+ (ResNet-101) \\
\midrule
\texttt{1} Baseline & 31.47 \\
\texttt{2} Baseline + \pst & \textbf{45.42} & {\color{blue}+13.95}\\
\midrule
SegFormer (MiT B5)\\
\midrule
\texttt{3} Baseline~\cite{hoyer2022daformer} & 45.60 \\
\texttt{4} Baseline + \pst & \textbf{52.57} & {\color{blue}+6.97}\\
\midrule
HRDA (MiT B5)\\
\midrule
\texttt{5} Baseline~\cite{hoyer2022hrda} & 53.01 \\
\texttt{6} Baseline + \pst & \textbf{57.21} & {\color{blue}+4.20}\\
\bottomrule
\end{tabular}}
\caption{\small\textbf{\pst is agnostic to choice of (SemSeg) architecture.} Semantic Segmentation models trained on GTAV (G) and evaluated on Cityscapes (C). 
\textbf{Bold} indicates best. $\Delta$ indicates (absolute) improvement over Baseline.
}
\label{table:semseg_cnn_transformer}
\end{center}
\vspace{-20pt}
\end{table}

%% file: sections/conclusion.tex
We propose Proportional Amplitude Spectrum Training Augmentation (\pst), a \textit{plug-and-play} augmentation strategy for synthetic-to-real generalization. 
\pst is motivated by the observation that the amplitude spectra are less diverse in synthetic than real data, especially for high-frequency components. 
Thus, \pst augments synthetic data by perturbing the amplitude spectra, with magnitudes increasing for higher frequencies.
We show that \pst offers strong out-of-the-box generalization performance on semantic segmentation, object detection, and object classification tasks. 
The strong performance of \pst holds true alone (i.e., training with ERM using \pst augmented images) or together with alternative generalization/augmentation algorithms. 
We would like to emphasize that the
strength of \pst lies in its simplicity (just modify your augmentation pipeline) and effectiveness, offering strong improvements despite not using extra modeling components, objectives, or 
data. We hope that future research endeavors in syn-to-real generalization take 
augmentation
techniques like \pst into account. Additionally, it might be of interest to the research community to explore how \pst could be utilized for \textit{adaptation} -- when ``little'' real target is available.
\looseness=-1

\newpage
\par \noindent
\textbf{Acknowledgments.} We thank Viraj Prabhu and George Stoica for fruitful discussions and valuable feedback. 
This work was supported in part by sponsorship from NSF Award \#2144194,
NASA ULI, ARL and Google.

%% file: appendix_sections/implementation_details.tex
In this section, we outline our training and implementation details for each of the three tasks -- Semantic Segmentation, Object Detection, and Object Recognition. We also summarize these details in Tables.~\ref{tab:segmentation_hparams}, \ref{tab:detection_hparams}, and~\ref{tab:recognition_hparams}.

\input{appendix_tables/hparam_details}

\par \noindent
\textbf{Semantic Segmentation (see Table.~\ref{tab:segmentation_hparams}).} 
For our primary semantic segmentation (SemSeg) experiments (in Tables 1, 4, and 6), we use the DeepLabv3+~\cite{chen2018encoder} architecture with backbones -- ResNet-50 (R-50)~\cite{he2016deep} and ResNet-101 (R-101)~\cite{he2016deep}. 
In Sec.~\ref{sec:syn2real}, we report additional results with DeepLabv3+ using the MobileNetv2 (Mn-v2)~\cite{sandler2018mobilenetv2} backbone.
We adopt the hyper-parameter (and distributed training) settings used by~\cite{choi2021robustnet} for training. 
Similar to~\cite{choi2021robustnet}, we train ResNet-50, ResNet-101 and MobileNet-v2 models in a distributed manner across 4, 4 and 2 GPUs respectively.
We use SGD (with momentum $0.9$) as the optimizer with an initial learning rate of $10^{-2}$ 
and
a polynomial learning rate schedule~\cite{liu2015parsenet} with a power of $0.9$. Our models are initialized with supervised ImageNet~\cite{krizhevsky2012imagenet} pre-trained weights. We train all our models for $40$k iterations with a batch size of $16$ for GTAV. Our segmentation models are trained on the train split of GTAV and evaluated on the validation splits of the target datasets (Cityscapes, BDD100K and Mapillary). For segmentation, \pst is applied with a base set of positional and photometric augmentations (\pst first and then the base augmentations) -- GaussianBlur, ColorJitter, RandomCrop, RandomHorizontalFlip and RandomScaling.
For RandAugment~\cite{cubuk2020randaugment}, we only consider the vocabulary of photometric augmentations for segmentation \& detection. We conduct ablations (within computational constraints) for the best performing RandAugment setting using R-50 for syn-to-real generalization and find that best performance is achieved when $8$ (photometric) augmentations are sampled at the highest severity level ($30$) from the augmentation vocabulary for application. Whenever we train a prior generalization approach, say ISW~\cite{choi2021robustnet} or IBN-Net~\cite{pan2018two}, we follow the same set of hyper-parameter configurations as
used in the respective papers.
Table.~\ref{tab:segmentation_hparams} includes details for SegFormer and HRDA runs.
All models except SegFormer and HRDA were trained across 3 random seeds.\looseness=-1

\par \noindent
\textbf{Object Detection (see Table.~\ref{tab:detection_hparams}).} For object detection (ObjDet), we use the Faster-RCNN~\cite{ren2015faster} architecture with ResNet-50 and ResNet-101 backbones (see 
Tables 2, 5 in the main paper).
Consistent with prior work~\cite{khindkar2022miss}, we train on the entirety of Sim10K~\cite{johnson2016driving} (source dataset) for $10$k iterations and pick the last checkpoint for Cityscapes (target dataset) evaluation. We use SGD with momentum as our optimizer with an initial learning rate of $10^{-2}$ (adjusted according to a step learning rate schedule) and a batch size of $32$. Our models are initialized with supervised ImageNet~\cite{krizhevsky2012imagenet} pre-trained weights. All models are trained on 4 GPUs in a distributed manner. For detection, we also compare \pst against RandAugment~\cite{cubuk2020randaugment} and Photometric Distortion (PD). The sequence of operations in PD to augment input images are -- randomized brightness, randomized contrast, RGB$\to$HSV conversion, randomized saturation \& hue changes, HSV$\to$RGB conversion, randomized contrast, and randomized channel swap.

\par \noindent
\textbf{Object Recognition (see Table.~\ref{tab:recognition_hparams}).} 
For our primary object recognition (ObjRec) experiments (see Table 3 in main paper), we train classifiers with ResNet-101~\cite{he2016deep} and ViT-B/16~\cite{dosovitskiy2021an} backbones. For ResNet-101, we start from supervised ImageNet~\cite{krizhevsky2012imagenet} pre-trained weights. For ViT-B/16 we start from both supervised and self-supervised (DINO~\cite{caron2021emerging}) ImageNet pre-trained weights. We train these classifiers for $10$ epochs with SGD (momentum $0.9$, weight decay $10^{-4}$) with an initial learning rate of $2\times10^{-4}$ with cosine annealing -- the newly added classifier and bottleneck layers~\cite{chen2022contrastive} were trained with $10\times$ more learning rate as the rest of the network. We train on $90\%$ of the (synthetic) VisDA-C training split (and use the remaining $10\%$ for model selection) with a batch size of 128 in a distributed manner across 4 GPUs. We use RandomCrop, RandomHorizontalFlip as additional augmentations with \pst. In Sec.~\ref{sec:syn2real}, we provide additional results demonstrating how \pst is complementary to CSG~\cite{chen2021contrastive}, a state-of-the-art generalization method on VisDA-C. For these experiments, to ensure fair comparisons, we train ResNet-101 based classifiers (with supervised ImageNet pre-trained weights) with same configurations as~\cite{chen2021contrastive}. This includes the use of an SGD (with momentum $0.9$) optimizer with a learning rate of $10^{-4}$, weight decay of $5\times10^{-4}$ and a batch size of $32$. These models are trained for $30$ epochs. CSG~\cite{chen2021contrastive} also uses RandAugment~\cite{cubuk2020randaugment} as an augmentation -- we check the effectiveness of \pst when applied with and without RandAugment during training.

%% file: appendix_tables/hparam_details.tex
\begin{table*}
\begin{subtable}{1\textwidth}
\caption{\small Semantic Segmentation Training}
\footnotesize
\centering
\begin{tabular}{l | l | l | l}
\textbf{Config} & \textbf{Value} (Tables~\ref{table:semseg_baseline_improvements},~\ref{table:semseg_sota} and~\ref{table:semseg_complementary}) & \textbf{Value} (Table.~\ref{table:semseg_cnn_transformer}; SegFormer) & \textbf{Value} (Table.~\ref{table:semseg_cnn_transformer}; HRDA)\\
\toprule
Source Data & GTAV (Train Split) & GTAV (Entirety) & GTAV (Entirety) \\
Target Data & Cityscapes (Val Split) & Cityscapes (Val Split) & Cityscapes (Val Split)\\
& BDD100K (Val Split) \\
& Mapillary (Val Split) \\
Segmentation Architecture & DeepLabv3+~\cite{chen2018encoder} & SegFormer~\cite{xie2021segformer} & HRDA~\cite{hoyer2022hrda}\\
Backbones & ResNet-50 (R-50)~\cite{he2016deep} & MiT-B5~\cite{xie2021segformer} & MiT-B5~\cite{xie2021segformer}\\
& ResNet-101 (R-101)~\cite{he2016deep}\\
& MobileNetv2 (Mn-v2)~\cite{sandler2018mobilenetv2} \\
Training Resolution & Original GTAV resolution & Original GTAV resolution & Original GTAV resolution\\
Optimizer & SGD & AdamW & AdamW\\
Initial Learning Rate & $10^{-2}$ & $6\times10^{-5}$ & $6\times10^{-5}$ \\
Learning Rate Schedule & Poly-LR  & Poly-LR & Poly-LR\\
Initialization & Imagenet Pre-trained Weights~\cite{krizhevsky2012imagenet} & Imagenet Pre-trained Weights~\cite{krizhevsky2012imagenet} & Imagenet Pre-trained Weights~\cite{krizhevsky2012imagenet} \\
Iterations & $40$k & $160$k & $40$k\\
Batch Size & $16$ & $4$ & $2$\\
Augmentations w/ \pst & Gaussian Blur, Color Jitter, Random Crop & Photometric Distortion & Photometric Distortion\\
& Random Horizontal Flip, Random Scaling & Random Crop, Random Flip & Random Crop, Random Flip \\
Model Selection Criteria & Best in-domain validation performance & End of training & End of training\\
GPUs & $4$ (R-50/101) or $2$ (Mn-v2) & $4$ & $1$ \\
\end{tabular}
\label{tab:segmentation_hparams} 
\vspace{1em}
\end{subtable}
\hfill \\
\begin{subtable}{1\textwidth}
\caption{\small Object Detection Training}
\footnotesize
\centering
\begin{tabular}{l | l}
\textbf{Config} & \textbf{Value} (Tables~\ref{table:detection_baseline_improvements} and~\ref{table:detection_sota}) \\
\toprule
Source Data & Sim10K \\
Target Data & Cityscapes (Val Split) \\
Segmentation Architecture & Faster-RCNN~\cite{ren2015faster} \\
CNN Backbones & ResNet-50 (R-50)~\cite{he2016deep} \\
& ResNet-101 (R-101)~\cite{he2016deep}\\
Training Resolution & Original Sim10K resolution for R-50, R-101 \\
Optimizer & SGD (momentum = $0.9$) \\
Initial Learning Rate & $10^{-2}$ \\
Learning Rate Schedule & Step-LR, Warmup $500$ iterations, Warmup Ratio $0.001$ \\
& Steps $6$k \& $8$k iterations \\
Initialization & Imagenet Pre-trained Weights~\cite{krizhevsky2012imagenet} \\
Iterations & $10$k \\
Batch Size & $32$ \\
Augmentations w/ \pst & Resize, Random Flip\\
Model Selection Criteria & End of training \\
GPUs & $4$ \\
\end{tabular}
\label{tab:detection_hparams} 
\vspace{1em}
\end{subtable}
\hfill \\
\begin{subtable}{1\textwidth}
\caption{\small Object Recognition Training}
\footnotesize
\centering
\begin{tabular}{l | l | l}
\textbf{Config} & \textbf{Value} (Table.~\ref{table:r101_visda_cls_res_baseline}) & \textbf{Value} (for Table.~\ref{table:r101_visda_cls_res}, following~\cite{chen2021contrastive}) \\
\toprule
Source Data & VisDA-C Synthetic & VisDA-C Synthetic \\
Target Data & VisDA-C Real & VisDA-C Real \\
Backbone & ResNet-101 (R-101)~\cite{he2016deep} & ResNet-101 (R-101)~\cite{he2016deep} \\
 & ViT-B/16~\cite{dosovitskiy2021an} (Sup \& DINO~\cite{caron2021emerging}) & \\
Optimizer & SGD w/ momentum (0.9) & SGD w/ momentum (0.9)\\
Initial Learning Rate & $2\times10^{-4}$ & $10^{-4}$  \\
Weight Decay & $10^{-4}$ & $5\times10^{-4}$\\
Initialization & Imagenet~\cite{krizhevsky2012imagenet} & Imagenet~\cite{krizhevsky2012imagenet}\\
Epochs & $10$ & $30$ \\
Batch Size & $128$ & $32$ \\
Augmentations w/ \pst & RandomCrop, RandomHorizontalFlip & RandAugment~\cite{cubuk2020randaugment} \\
Model Selection Criteria & Best in-domain val performance & Best in-domain val performance \\
GPUs & $1$ (CNN) or $4$ (ViT) & $1$ \\
\end{tabular}
\vspace{-.5em}
\label{tab:recognition_hparams} 
\end{subtable}
\caption{\small\textbf{Implementation \& Optimization Details.} We summarize details surrounding dataset, training, optimization and model selection criteria for our semantic segmentation, object detection and object recognition experiments. More detailed configs in code.}
\label{tab:pt_hp}
\vspace{20pt}
\end{table*}

%% file: appendix_sections/syn_to_real_results.tex
\input{appendix_tables/mnv2_seg_res_fullres_train}

\par \noindent
\textbf{MobileNet-v2 GTAV$\to$Real Generalization Results.} Our key generalization results for semantic segmentation (SemSeg) (in Tables. 1, 4 and 6) are with ResNet-50 and ResNet-101 backbones. In Table.~\ref{table:mnv2_semseg_results}, we also report results when \pst is applied to DeepLabv3+~\cite{chen2018encoder} models with MobileNetv2 -- a lighter backbone tailored for resource constrained settings. We find that \pst substantially improves a vanilla baseline (by $14+$ absolute mIoU points; rows 1, 2) and is complementary to existing methods (rows 3-6).

\par \noindent
\textbf{Prior work using more data / different architectures.} In Table. 4
of the main paper, 
we compare models trained using \pst with prior syn-to-real generalization methods.
As stated in Sec. 5.1 of the main paper, 
prior methods such as DRPC~\cite{yue2019domain}, ASG~\cite{chen2020automated}, CSG~\cite{chen2021contrastive},  WEDGE~\cite{kim2023wedge}, FSDR~\cite{huang2021fsdr} and WildNet~\cite{lee2022wildnet} \& SHADE~\cite{zhao2022shade} (for R-101) use either more data or different base architectures, making these comparisons unfair \textit{to} \pst.
For instance, WEDGE~\cite{kim2023wedge} and CSG~\cite{chen2021contrastive} use DeepLabv2, ASG~\cite{chen2020automated} 
uses FCNs, DRPC~\cite{yue2019domain} \& SHADE~\cite{zhao2022shade} (for R-101)
use the entirety of GTAV (not just the training split; $2\times$ more data compared to \pst) and WEDGE uses $\sim$5k extra Flickr images in it's overall pipeline. 
FSDR~\cite{huang2021fsdr} uses FCNs and the the entirety of GTAV for training.
FSDR~\cite{huang2021fsdr} and WildNet~\cite{lee2022wildnet} also use extra ImageNet~\cite{krizhevsky2012imagenet} images for stylization / randomization. For FSDR, the first step in the pipeline also requires access to SYNTHIA~\cite{ros2016synthia}. Unlike \pst, FSDR and DRPC also select the checkpoints that perform best on target data~\cite{fsdr_issue, yue2019domain}, which is unrealistic since assuming access to labeled (real) target data (for training or model selection) is not practical for syn-to-real generalization. We note that despite having access to \textit{less} data, \pst outperforms these methods on GTAV$\to$Real.

\input{appendix_tables/recognition_pasta_complementary}

\par \noindent
\textbf{\pst complementary to CSG~\cite{chen2021contrastive}.} To evaluate the efficacy of \pst for object recognition (ObjRec), in Table.~\ref{table:r101_visda_cls_res}, we apply \pst to CSG~\cite{chen2021contrastive}, a state-of-the-art generalization method on VisDA-C Syn$\to$Real. Since CSG inherently uses RandAugment~\cite{cubuk2020randaugment}, we apply \pst $(\alpha=10,k=1,\beta=0.5)$ both with (row 6) and without (row 5) RandAugment and find that applying \pst improves over vanilla CSG (row 4) in both conditions.

\input{appendix_tables/gtav_2_cityscapes_per_cls_res_unresized}

\input{appendix_tables/gtav_2_bdd_per_cls_res_unresized}

\input{appendix_tables/gtav_2_mapillary_per_cls_res_unresized}

\par \noindent
\textbf{Per-class GTAV$\to$Real Generalization Results.} Tables~\ref{table:gtav_2_cityscapes_r50_per_cls_res_unresized},~\ref{table:gtav_2_bdd_r50_per_cls_res_unresized} and~\ref{table:gtav_2_mapillary_r50_per_cls_res_unresized} include per-class synthetic-to-real generalization results when a DeepLabv3+ (R-50 backbone) model trained on GTAV is evaluated on Cityscapes, BDD100K and Mapillary respectively. 
For GTAV$\to$Cityscapes (see Table.~\ref{table:gtav_2_cityscapes_r50_per_cls_res_unresized}), we find that Baseline + \pst consistently improves over Baseline and RandAugment. For IBN-Net and ISW in this setting, we observe consistent improvements (except for the classes \textit{terrain} and \textit{fence}).
For GTAV$\to$BDD100K (see Table.~\ref{table:gtav_2_bdd_r50_per_cls_res_unresized}), we find that for the Baseline, while \pst outperforms RandAugment on the majority of classes, both are fairly competitive and outperform the vanilla Baseline approach. For IBN-Net and ISW, \pst 
almost always outperforms the vanilla approaches (except for the class \textit{wall}). For GTAV$\to$Mapillary (see Table.~\ref{table:gtav_2_mapillary_r50_per_cls_res_unresized}), for Baseline, we find that \pst 
outperforms the vanilla approach and RandAugment.
For IBN-Net and ISW, \pst 
outperforms the vanilla approaches with the exception of the classes \textit{train} and \textit{fence}.

\par \noindent
\textbf{\pst helps SYNTHIA$\to$Real Generalization.}
We conducted additional syn-to-real experiments using SYNTHIA~\cite{ros2016synthia} as the source domain and Cityscapes, BDD100K and Mapillary as the target domains.
For a baseline DeepLabv3+ model (R-101), we find that \pst – (1) provides strong improvements over the vanilla baseline ($31.77$\% mIoU, $+3.91$\% absolute improvement) and (2) is competitive with RandAugment ($32.30$\% mIoU). More generally, we find that syn-to-real generalization performance is worse when SYNTHIA is used as the source domain as opposed to GTAV – for instance, ISW~\cite{choi2021robustnet} achieves an average mIoU of $31.07$\% (SYNTHIA) as opposed to $35.58$\% (GTAV). SYNTHIA has significantly fewer images compared to GTAV ($9.4$k vs $25$k), which likely contributes to relatively worse generalization performance.

\input{appendix_tables/pasta_base_augmentations}

\par \noindent
\textbf{\pst and Base Augmentations.} 
\pst is applied with some consistent color and positional augmentations (see Section.~\ref{sec:implementation}). To understand if \pst alone leads to any improvements, in Table.~\ref{table:pasta_base_augs}, we conduct a controlled experiments where we train a baseline DeepLabv3+ model (R-50) on GTAV (by downsampling input images to a resolution of $1024\times560$ due to compute constraints) with different augmentations and evaluate on real data (Cityscapes, BDD100K and Mapillary). We find that applying \pst alone leads to significant improvements ($13+$ absolute mIoU points; row 2) and including the positional (row 3) and photometric (row 4) augmentations leads to further improvements.

%% file: appendix_tables/mnv2_seg_res_fullres_train.tex
\begin{table}[t]
\footnotesize
\centering
\setlength{\tabcolsep}{3.0pt}
\begin{center}
\resizebox{\columnwidth}{!}{
\begin{tabular}{lccccc}
\toprule
\multirow{2}{*}{\textbf{Method}} & \multicolumn{4}{c}{\textbf{Real mIoU} $\uparrow$}\\ \cmidrule{2-6}
 & G$\rightarrow$C &G$\rightarrow$B & G$\rightarrow$M & Avg & $\Delta$\\
\midrule
\multicolumn{1}{l}{\texttt{1} Baseline (B)~\cite{choi2021robustnet}$^*$} & 25.92 & 25.73 & 26.45 & 26.03 \\
\multicolumn{1}{l}{\texttt{2} B + \pst} & \textbf{39.75} & \textbf{37.54} & \textbf{43.28} &  \textbf{40.19\tiny{\std{0.45}}} & {\color{blue}+14.16}\\
\midrule
\multicolumn{1}{l}{\texttt{3} IBN-Net~\cite{pan2018two}$^*$} & 30.14 & 27.66 & 27.07 & 28.29 \\
\multicolumn{1}{l}{\texttt{4} IBN-Net + \pst} & \textbf{37.57} & \textbf{36.97} & \textbf{40.91} &  \textbf{38.48\tiny{\std{0.75}}} & {\color{blue}+10.19}\\
\midrule
\multicolumn{1}{l}{\texttt{5} ISW~\cite{choi2021robustnet}$^*$} & 30.86 & 30.05 & 30.67 & 30.53 \\
\multicolumn{1}{l}{\texttt{6} ISW + \pst} & \textbf{37.99} & \textbf{37.49} & \textbf{42.44} &  \textbf{39.31\tiny{\std{1.26}}} & {\color{blue}+8.79}\\
\bottomrule
\end{tabular}}
\caption{\small\textbf{MobileNet-v2~\cite{sandler2018mobilenetv2} GTAV$\to$Real (SemSeg) Generalization Results.} Semantic Segmentation DeepLabv3+ models trained on GTAV (G) and evaluated on $\{$Cityscapes (C), BDD100K (B), Mapillary (M)$\}$. $^*$ indicates numbers drawn from published manuscripts. 
\textbf{Bold} indicates best. $\Delta$ indicates (absolute) improvement.
\pst improves a vanilla baseline (rows 1, 2) and is complementary to existing methods (rows 3-6).}
\label{table:mnv2_semseg_results}
\end{center}
\vspace{-15pt}
\end{table}

%% file: appendix_tables/recognition_pasta_complementary.tex
\begin{table}[t]
\footnotesize
\centering
\setlength{\tabcolsep}{2.5pt}
\begin{center}
\resizebox{0.85\columnwidth}{!}{
\begin{tabular}{lcc}
\toprule
\textbf{Method} & \textbf{Accuracy} & $\Delta$ \\
\midrule
\texttt{1} Oracle (IN-1k)~\cite{chen2021contrastive}$^*$ & 53.30 \\
\texttt{2} Baseline (Syn. Training)~\cite{chen2021contrastive}$^*$ & 49.30 \\
\texttt{3} CSG~\cite{chen2021contrastive}$^*$ & 64.05 \\
\texttt{4} CSG (RandAug) & 63.84\std{0.29} & {\color{blue} +14.54}\\
\texttt{5} CSG (\pst)& 64.29\std{0.56} & {\color{blue} +14.99}\\
\texttt{6} CSG (RandAug + \pst)& \textbf{65.86\std{1.13}} & {\color{blue} +16.56}\\
\bottomrule
\end{tabular}}
\caption{\small\textbf{\pst is complementary to CSG~\cite{chen2021contrastive}.} We apply \pst to CSG~\cite{chen2021contrastive}. Since CSG inherently uses RandAug, we also report results with and without the use of RandAug when \pst is applied. $^*$ indicates drawn directly from published manuscripts. We report class-balanced accuracy on the real (val split) target data of VisDA-C. 
Results reported across 3 runs. \textbf{Bold} indicates best. $\Delta$ indicates absolute improvement over baseline.}
\label{table:r101_visda_cls_res}
\end{center}
\vspace{-20pt}
\end{table}

%% file: appendix_tables/gtav_2_cityscapes_per_cls_res_unresized.tex
\begin{table*}[t]
    \setlength{\tabcolsep}{2pt}
    \begin{center}  
    \resizebox{\textwidth}{!}{
    \begin{tabular}{lcccccccccccccccccccc}
    \toprule
    \textbf{Method} &  \rotb{road} & \rotb{building} & \rotb{vegetation} & \rotb{car} & \rotb{sidewalk} & \rotb{sky} & \rotb{pole} & \rotb{person} & \rotb{terrain} & \rotb{fence} & \rotb{wall} & \rotb{bicycle} & \rotb{sign} & \rotb{bus} & \rotb{truck} & \rotb{rider} & \rotb{light} & \rotb{train} & \rotb{motorcycle} &  \textbf{mIoU}  \\
    \midrule
    \texttt{1} Baseline (B)~\cite{choi2021robustnet}$^*$ & 45.1 & 56.8 & 80.9 & 61.0 & 23.1 & 38.9 & 23.9 & 58.2 & 24.3 & 16.3 & 16.6 & 13.4 & 7.3 & 20.0 & 17.4 & 1.2 & 30.0 & 7.2 & 8.5 & 29.0\\
    \texttt{2} B + RandAug & 58.5 & 56.3 & 77.3 & 83.7 & 30.3 & 45.1 & 27.3 & 57.8 & 20.6 & 20.9 & 11.4 & 16.9 & 9.7 & 20.3 & 15.0 & 2.4 & 28.1 & 14.0 & 10.3 & 31.9\\
    \texttt{3} B + \pst & \textbf{84.1} & \textbf{80.5} & \textbf{85.8} & \textbf{85.9} & \textbf{40.1} & \textbf{81.8} & \textbf{31.9} & \textbf{66.0} & \textbf{31.4} & \textbf{28.1} & \textbf{29.0} & \textbf{21.8} & \textbf{28.5} & \textbf{24.5} & \textbf{28.7} & \textbf{7.0} & \textbf{32.9} & \textbf{23.4} & \textbf{27.2} & \textbf{44.1}\\
    \midrule
    \texttt{4} IBN-Net~\cite{pan2018two}$^*$ & 51.3 & 59.7 & 85.0 & 76.7 & 24.1 & 67.8 & 23.0 & 60.6 & \textbf{40.6} & \textbf{25.9} & 14.1 & 15.7 & 10.1 & 23.7 & 16.3 & 0.8 & 30.9 & 4.9 & 11.9 & 33.9\\
    \texttt{5} IBN-Net + \pst & \textbf{78.1} & \textbf{79.5} & \textbf{85.8} & \textbf{84.5} & \textbf{31.7} & \textbf{80.1} & \textbf{32.2} & \textbf{63.4} & 38.8 & 21.7 & \textbf{28.0} & \textbf{18.2} & \textbf{22.6} & \textbf{26.4} & \textbf{29.0} & \textbf{2.8} & \textbf{34.0} & \textbf{16.5} & \textbf{22.9} & \textbf{41.9}\\
    \midrule
    \texttt{6} ISW~\cite{choi2021robustnet}$^*$ & 60.5 & 65.4 & 85.4 & 82.7 & 25.5 & 70.3 & 25.8 & 61.9 & 38.5 & \textbf{23.7} & 21.6 & 15.5 & 12.2 & 25.4 & 21.1 & 0.0 & 33.3 & 9.3 & 16.8 & 36.6\\
    \texttt{7} ISW + \pst & \textbf{76.6} & \textbf{78.4} & \textbf{85.6} & \textbf{83.7} & \textbf{32.5} & \textbf{83.1} & \textbf{33.1} & \textbf{63.4} & \textbf{40.4} & 23.6 & \textbf{27.3} & \textbf{17.4} & \textbf{22.3} & \textbf{25.7} & \textbf{30.1} & \textbf{3.3} & \textbf{35.9} & \textbf{18.2} & \textbf{19.9} & \textbf{42.1}\\
    \bottomrule
    \end{tabular}}
    \caption{\small \textbf{GTAV$\to$Cityscapes per-class generalization results.} Per-class IoU comparisons for (SemSeg) syn-to-real generalization results when DeepLabv3+ (R-50 models trained on GTAV are evaluated on Cityscapes. Results are reported across 3 runs. $^*$ indicates drawn directly from published manuscripts. 
    Class headers are in decreasing order of pixel frequency.
    }
    \label{table:gtav_2_cityscapes_r50_per_cls_res_unresized}
    \end{center}
\end{table*}

%% file: appendix_tables/gtav_2_bdd_per_cls_res_unresized.tex
\begin{table*}[ht!]    
    \setlength{\tabcolsep}{2pt}
    \begin{center}
    \resizebox{\textwidth}{!}{
    \begin{tabular}{lcccccccccccccccccccc}
    \toprule
    \textbf{Method} &  \rotb{road} & \rotb{sky} & \rotb{building} & \rotb{vegetation} & \rotb{car} & \rotb{sidewalk} & \rotb{fence} & \rotb{terrain} & \rotb{truck} & \rotb{pole} & \rotb{bus} & \rotb{wall} & \rotb{sign} & \rotb{person} & \rotb{light} & \rotb{bicycle} & \rotb{motorcycle} & \rotb{rider} & \rotb{train} &  \textbf{mIoU}  \\
    \midrule
    \texttt{1} Baseline (B)~\cite{choi2021robustnet}$^*$ & 48.2 & 32.3 & 34.3 & 58.2 & 67.3 & 23.0 & 19.8 & 21.4 & 11.3 & 28.4 & 5.6 & 3.4 & 18.1 & 43.9 & 30.2 & 11.1 & 16.0 & 5.1 & 0.0 & 25.1 \\
    \texttt{2} B + RandAug & 75.4 & 82.8 & 67.7 & \textbf{74.7} & 74.1 & \textbf{39.3} & \textbf{32.7} & \textbf{26.6} & 22.7 & \textbf{37.0} & \textbf{16.5} & 5.0 & 23.9 & 51.7 & 35.8 & 12.0 & 29.3 & \textbf{20.2} & 0.0 & 38.3\\
    \texttt{3} B + \pst & \textbf{86.0} & \textbf{86.6} & \textbf{74.8} & 72.7 & \textbf{82.8} & 38.4 & 31.2 & 24.9 & \textbf{23.7} & 34.8 & 6.4 & \textbf{11.2} & \textbf{26.2} & \textbf{55.1} & \textbf{37.0} & \textbf{13.3} & \textbf{38.3} & 19.9 & 0.0 & \textbf{40.2}\\
    \midrule
    \texttt{4} IBN-Net~\cite{pan2018two}$^*$ & 68.9 & 66.9 & 56.7 & 66.6 & 70.3 & 28.8 & 21.4 & 22.1 & 12.8 & 31.9 & 7.2 & 6.0 & 21.7 & 50.2 & 35.0 & 18.1 & 23.2 & 5.8 & 0.0 & 32.3\\
    \texttt{5} IBN-Net + \pst & \textbf{86.1} & \textbf{87.6} & \textbf{74.9} & \textbf{72.3} & \textbf{82.3} & \textbf{36.6} & \textbf{30.6} & \textbf{26.2} & \textbf{25.3} & \textbf{37.1} & \textbf{10.8} & \textbf{13.2} & \textbf{25.5} & \textbf{56.0} & \textbf{36.8} & \textbf{21.4} & \textbf{38.9} & \textbf{26.0} & 0.0 & \textbf{41.5}\\
    \midrule
    \texttt{6} ISW~\cite{choi2021robustnet}$^*$ & 74.9 & 77.4 & 65.2 & 69.0 & 72.4 & 30.4 & 22.6 & 26.2 & 16.2 & 34.9 & 6.1 & \textbf{11.5} & 22.2 & 50.3 & 36.9 & 11.4 & 31.3 & 10.0 & 0.0 & 35.2\\
    \texttt{7} ISW + \pst & \textbf{86.5} & \textbf{87.9} & \textbf{74.0} & \textbf{73.0} & \textbf{83.2} & \textbf{37.7} & \textbf{28.6} & \textbf{28.1} & \textbf{23.4} & \textbf{37.2} & \textbf{7.8} & 11.3 & \textbf{25.0} & \textbf{55.1} & \textbf{37.8} & \textbf{23.6} & \textbf{35.5} & \textbf{22.4} & 0.0 & \textbf{41.0}\\
    \bottomrule
    \end{tabular}}
    \caption{\small \textbf{GTAV$\to$BDD100K per-class generalization results.} Per-class IoU comparisons for (SemSeg) syn-to-real generalization results when DeepLabv3+ (R-50 models trained on GTAV are evaluated on BDD100K. Results are reported across 3 runs. $^*$ indicates drawn directly from published manuscripts. 
    Class headers are in decreasing order of pixel frequency.
    }
    \label{table:gtav_2_bdd_r50_per_cls_res_unresized}
    \end{center}
\end{table*}

%% file: appendix_tables/gtav_2_mapillary_per_cls_res_unresized.tex
\begin{table*}[ht!]
    \setlength{\tabcolsep}{2pt}
    \begin{center}
    \resizebox{\textwidth}{!}{
    \begin{tabular}{lcccccccccccccccccccc}
    \toprule
    \textbf{Method} & \rotb{sky} & \rotb{road} & \rotb{vegetation} & \rotb{building} & \rotb{sidewalk} & \rotb{car} & \rotb{fence} & \rotb{pole} & \rotb{terrain} & \rotb{wall} & \rotb{sign} & \rotb{truck} & \rotb{person} & \rotb{bus} & \rotb{light} & \rotb{bicycle} & \rotb{rider} & \rotb{motorcycle} & \rotb{train} &  \textbf{mIoU}  \\
    \midrule
    \texttt{1} Baseline (B)~\cite{choi2021robustnet}$^*$ & 42.2 & 46.8 & 64.9 & 33.5 & 24.9 & 72.3 & 14.4 & 27.7 & 23.8 & 6.7 & 8.5 & 23.7 & 53.7 & 7.0 & 35.8 & 18.4 & 4.9 & 15.5 & 10.8 & 28.2\\
    \texttt{2} B + RandAug~\cite{cubuk2020randaugment} & 51.7 & 59.6 & 75.5 & 39.5 & 33.9 & 81.3 & 22.6 & 37.2 & 24.6 & 4.2 & 32.4 & 31.2 & 56.5 & 13.4 & 36.2 & 18.0 & 11.9 & 21.4 & 5.0 & 34.5\\
    \texttt{3} B + \pst & \textbf{93.3} & \textbf{83.0} & \textbf{76.3} & \textbf{76.9} & \textbf{40.2} & \textbf{83.3} & \textbf{27.1} & \textbf{40.9} & \textbf{37.1} & \textbf{19.2} & \textbf{50.4} & \textbf{35.2} & \textbf{63.3} & \textbf{19.5} & \textbf{41.4} & \textbf{29.4} & \textbf{25.2} & \textbf{38.1} & \textbf{15.2} & \textbf{47.1}\\
    \midrule
    \texttt{4} IBN-Net~\cite{pan2018two}$^*$ & 82.0 & 66.4 & 73.5 & 57.1 & 32.9 & 73.1 & 24.9 & 31.5 & 28.4 & 10.5 & 38.9 & 30.7 & 56.4 & 16.0 & 38.0 & 18.6 & 9.1 & 16.6 & \textbf{12.6} & 37.7\\
    \texttt{5} IBN-Net + \pst & \textbf{94.4} & \textbf{81.7} & \textbf{76.1} & \textbf{76.9} & \textbf{40.4} & \textbf{80.8} & \textbf{27.1} & \textbf{40.3} & \textbf{38.7} & \textbf{19.0} & \textbf{43.2} & \textbf{38.0} & \textbf{62.0} & \textbf{20.5} & \textbf{39.3} & \textbf{25.7} & \textbf{23.6} & \textbf{31.6} & 12.4 & \textbf{45.9}\\
    \midrule
    \texttt{6} ISW~\cite{choi2021robustnet}$^*$ & 88.2 & 74.8 & 74.3 & 66.1 & 36.2 & 78.7 & \textbf{26.0} & 35.4 & 30.2 & 15.2 & 36.6 & 33.3 & 58.6 & 14.4 & 37.9 & 17.8 & 11.1 & 20.4 & 11.0 & 40.3\\
    \texttt{7} ISW + \pst & \textbf{94.6} & \textbf{82.9} & \textbf{76.7} & \textbf{76.0} & \textbf{41.9} & \textbf{81.8} & 25.8 & \textbf{40.4} & \textbf{40.9} & \textbf{18.8} & \textbf{43.1} & \textbf{34.1} & \textbf{61.6} & \textbf{19.9} & \textbf{40.2} & \textbf{24.5} & \textbf{22.3} & \textbf{30.5} & \textbf{11.6} & \textbf{45.7}\\
    \bottomrule
    \end{tabular}}
    \caption{\small \textbf{GTAV$\to$Mapillary per-class generalization results.} Per-class IoU comparisons for (SemSeg) syn-to-real generalization results when DeepLabv3+ (R-50 models trained on GTAV are evaluated on Mapillary. Results are reported across 3 runs. $^*$ indicates drawn directly from published manuscripts. 
    Class headers are in decreasing order of pixel frequency.
    }
    \label{table:gtav_2_mapillary_r50_per_cls_res_unresized}
    \end{center}
    \vspace{10pt}
\end{table*}

%% file: appendix_tables/pasta_base_augmentations.tex
\begin{table}[t]
\footnotesize
\centering
\setlength{\tabcolsep}{2.5pt}
\begin{center}
\resizebox{\columnwidth}{!}{
\begin{tabular}{lcccc}
\toprule
\multirow{2}{*}{\textbf{Method}} & \multicolumn{2}{c}{\textbf{Base Augmentations}} & \multirow{2}{*}{\textbf{Real mIoU}} & \multirow{2}{*}{$\Delta$}\\
& Positional & Photometric & & \\
\midrule
\texttt{1} Baseline (B) & \cmark & \cmark & 26.99 & \\
\texttt{2} B + \pst & \xmark & \xmark & 40.25 & {\color{blue}+13.26}\\
\texttt{3} B + \pst & \cmark & \xmark & 40.37 & {\color{blue}+13.38}\\
\texttt{4} B + \pst & \cmark & \cmark & \textbf{41.90} & {\color{blue}+14.91}\\
\bottomrule
\end{tabular}}
\caption{\small\textbf{\pst vs Base Augmentations.} Semantic Segmentation DeepLabv3+ (R-50) models trained on GTAV (at an input resolution of $1024\times560$ due to compute constraints) and evaluated on $\{$Cityscapes, BDD100K, Mapillary$\}$. 
\textbf{Bold} indicates best. $\Delta$ indicates (absolute) improvement over Baseline.}
\label{table:pasta_base_augs}
\end{center}
\vspace{-20pt}
\end{table}

%% file: appendix_sections/pasta_analysis.tex
In this section, we provide more discussions surrounding different aspects of \pst.

\input{appendix_tables/pasta_freq_augmentations}

\par \noindent
\textbf{\pst vs other frequency domain augmentation methods.} As noted in Sec 5.2 of the main paper, prior work has also considered augmenting images in the frequency domain as part of their training pipeline. Notably, FSDR~\cite{huang2021fsdr} augments images in the fourier domain. Table. 4 in the main paper shows how \pst (when applied to a vanilla baseline) already outperforms FSDR (by $2.29$ absolute mIoU points). In Table.~\ref{table:pasta_fdom_augs}, we compare \pst with other frequency domain augmentation strategies (summarized below) when applied to a baseline DeepLabv3+ (R-50) SemSeg model trained on GTAV (by downsampling input images to a resolution of $1024\times560$ due to compute constraints) and evaluated on real data (Cityscapes, BDD100K and Mapillary).
\begin{packed_itemize}
\item \textbf{FDA~\cite{yang2020fda} -- }FDA is a recent approach for syn-to-real domain adaptation and naturally requires access to unlabeled target data. In FDA, to augment source images, low frequency bands of the amplitude spectra of source images are replaced with those of target -- essentially mimicking a cheap style transfer operation. Since we do not assume access to target data in our experimental settings, a direct comparison is not possible. Instead, we consider a proxy task where we intend to generalize to real datasets (Cityscapes, BDD100K, Mapillary) by assuming additional access to 6 real world street view images under different weather conditions (for style transfer) – sunny day, rainy day, cloudy day, etc. – in addition to synthetic images from GTAV. We find that \pst outperforms FDA (row 2 vs row 6 in Table.~\ref{table:pasta_fdom_augs}).
\item \textbf{APR-P~\cite{chen2021amplitude} -- }Amplitude Phase Recombination is a recent method designed to improve robustness against natural corruptions. APR replaces the amplitude spectrum of an image with the amplitude spectrum from an augmented view (APR-S) or different images (APR-P). When applied to synthetic images from GTAV, we find that \pst outperforms APR-P (row 3 vs row 6 in Table.~\ref{table:pasta_fdom_augs}).
\item \textbf{FACT~\cite{xu2021fourier} -- }FACT is a 
multi-source domain generalization method for object recognition that uses one of two frequency domain augmentation strategies -- Amplitude Jitter (AJ) and Amplitude Mixup (AM) -- in a broader training pipeline. AM involves perturbing the amplitude spectrum of the image of concern by performing a convex combination with the amplitude spectrum of another ``mixup'' image drawn from the same source data. AJ (AJ) perturbs the amplitude spectrum with a single jitter value $\epsilon$ for all spatial frequencies and channels. We compare with both AM and AJ for SemSeg and find that \pst outperforms both (rows 4, 5 vs row 6 in Table.~\ref{table:pasta_fdom_augs}). Additionally, for multi-source domain generalization on PACS~\cite{li2017deeper}, we find that FACT-\pst (where we replace the augmentations in FACT with \pst) outperforms FACT-Vanilla -- $87.97\%$ vs $87.10\%$ average leave-one-out domain accuracy.\looseness=-1
\end{packed_itemize}

%% file: appendix_tables/pasta_freq_augmentations.tex
\begin{table}[t]
\footnotesize
\centering
\setlength{\tabcolsep}{2.5pt}
\begin{center}
\resizebox{\columnwidth}{!}{
\begin{tabular}{lccc}
\toprule
\textbf{Method} & \textbf{Real Data} & \textbf{Real mIoU} & $\Delta$ \\
\midrule
\texttt{1} Baseline & \xmark  & 26.99 \\
\rowcolor{Gray}
\texttt{2} Baseline + FDA~\cite{yang2020fda} & \cmark & 33.04 & {\color{blue}+6.05}\\
\texttt{3} Baseline + APR-P~\cite{chen2021amplitude} & \xmark & 37.52 & {\color{blue}+10.53}\\
\texttt{4} Baseline + AJ (FACT~\cite{xu2021fourier}) & \xmark & 30.70 & {\color{blue}+3.71}\\
\texttt{5} Baseline + AM (FACT~\cite{xu2021fourier}) & \xmark & 39.70 & {\color{blue}+12.71}\\
\texttt{6} Baseline + \pst & \xmark & 41.90 & {\color{blue}+14.91}\\
\bottomrule
\end{tabular}}
\caption{\small\textbf{\pst vs Frequency-domain Augmentations.} Semantic Segmentation DeepLabv3+ (R-50) models trained with different frequency domain augmentation strategies on GTAV (at an input resolution of $1024\times560$ due to compute constraints) and evaluated on $\{$Cityscapes, BDD100K, Mapillary$\}$. 
\textbf{Bold} indicates best. $\Delta$ indicates (absolute) improvement over Baseline. Row in \unfair{gray} uses real data for augmenting images.}
\label{table:pasta_fdom_augs}
\end{center}
\vspace{-20pt}
\end{table}

%% file: appendix_sections/amplitude_analysis.tex
\pst 
relies on
the empirical observation that synthetic images have less variance in their high frequency components compared to real images. In this section, we first show how this observation is widespread across a set of syn-to-real shifts over fine-grained frequency band discretizations and then demonstrate how \pst helps counter this discrepancy.

\par
\begin{figure*}
\centering
\includegraphics[width=\textwidth]{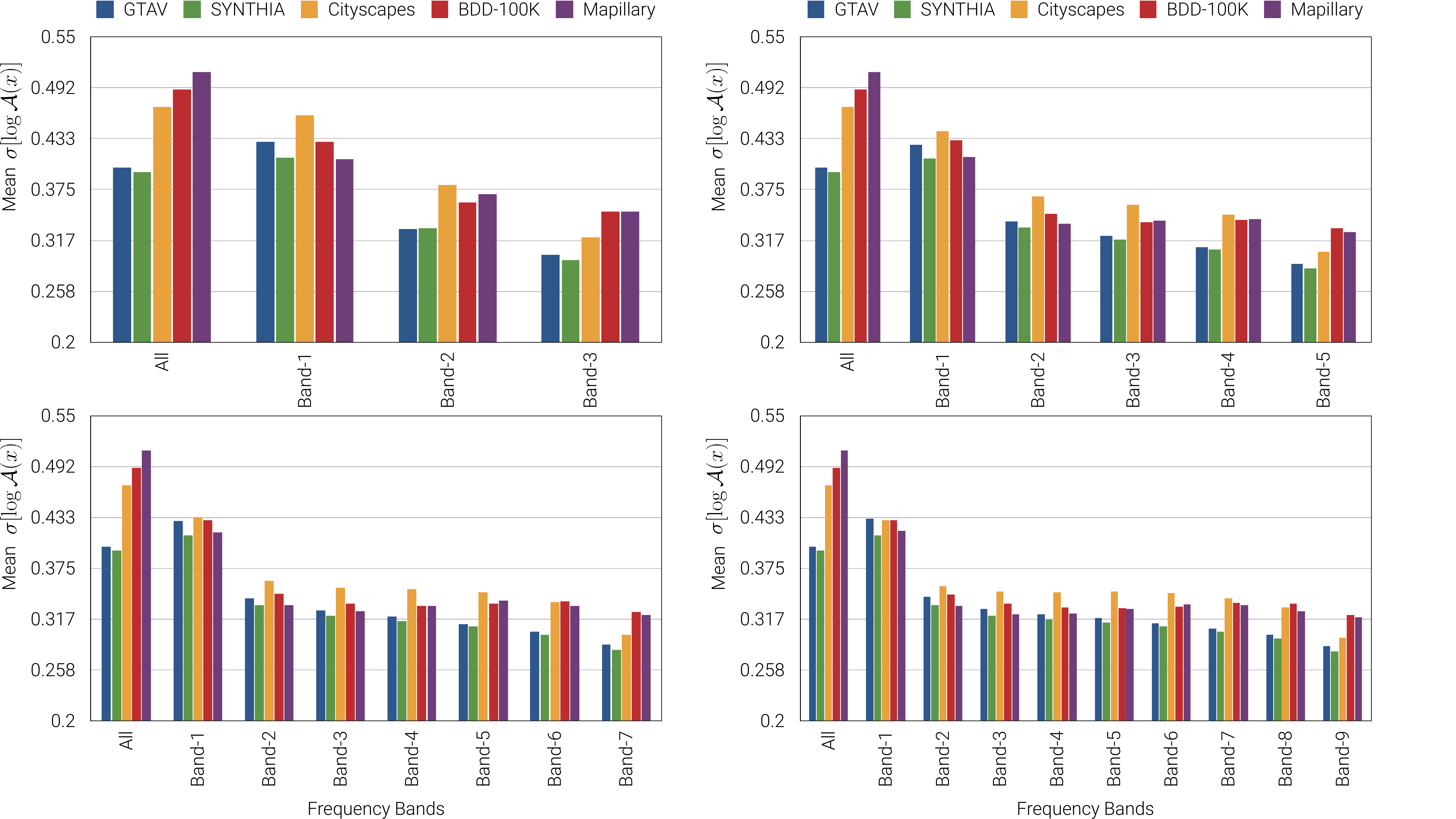}
\caption{\textbf{Variations in amplitude values across fine-grained frequency bands (GTAV$\to$Real and SYNTHIA$\to$Real).} Across domain shifts GTAV$\to$Real and SYNTHIA$\to$Real, and four settings corresponding to fine-grained frequency bands (3, 5, 7 and 9 bands; increasing in frequency from Band-1 to Band-$n$), we find that synthetic images have less variance in high-frequency components of the amplitude spectrum compared to real images.}
\label{fig:motivation_analysis_1}
\end{figure*}

\begin{figure*}
\centering
\includegraphics[width=\textwidth]{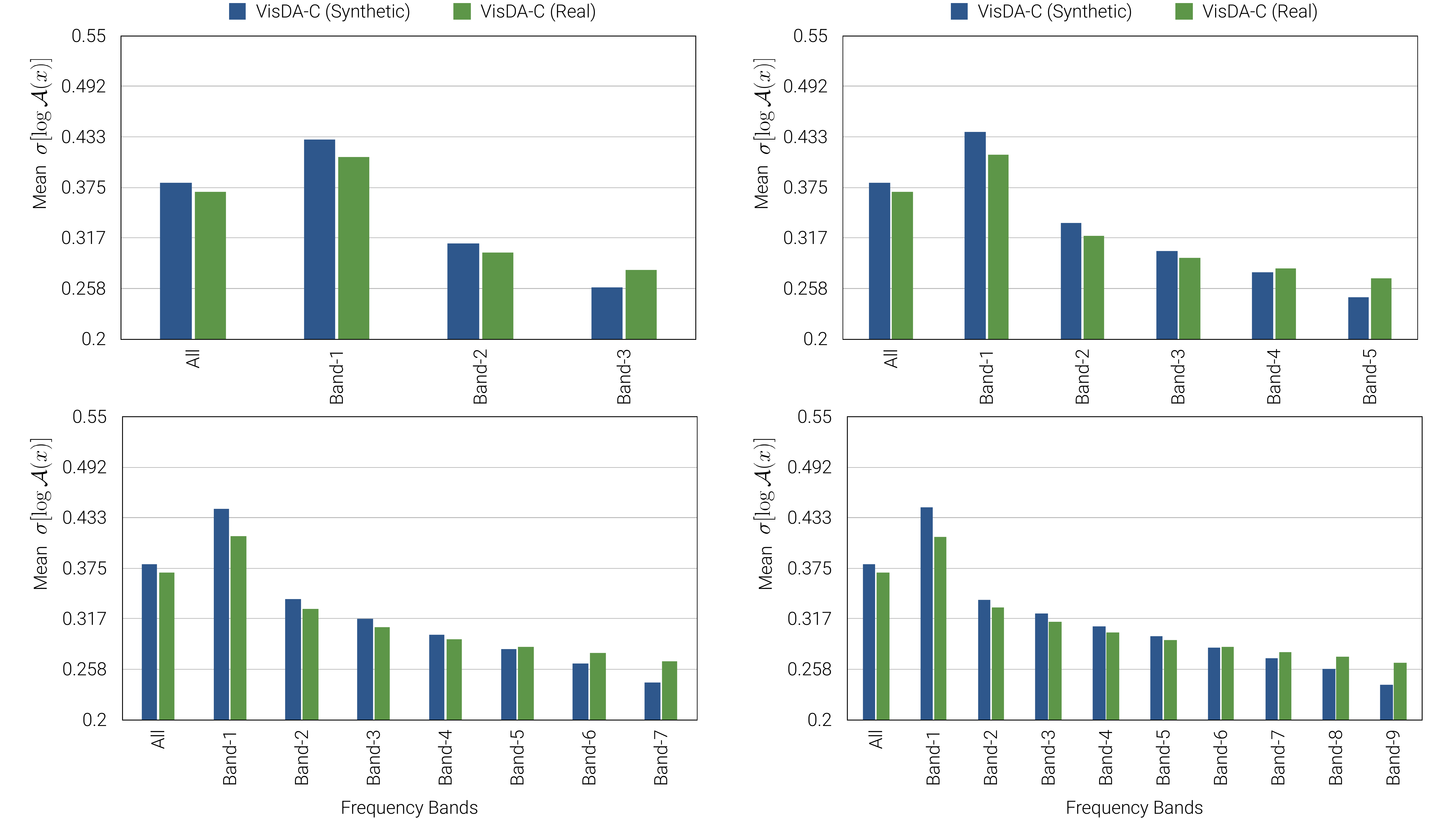}
\caption{\textbf{Variations in amplitude values across fine-grained frequency bands (VisDA-C Synthetic$\to$Real).} For the VisDA-C Synthetic$\to$Real domain shift, and four settings corresponding to fine-grained frequency bands (3, 5, 7 and 9 bands; increasing in frequency from Band-1 to Band-$n$), we find that synthetic images have less variance in high-frequency components of the amplitude spectrum compared to real images.}
\label{fig:motivation_analysis_2}
\end{figure*}

 \noindent
\textbf{Fine-grained Band Discretization.} For Fig. 2 [Right] in the main paper, the low, mid and high frequency bands are chosen such that the first (lowest) band is $1/3$ the height of the image (includes all spatial frequencies until $1/3$ of the image height), second band is up to $2/3$ the height of the image excluding band 1 frequencies, and the third band considers all the remaining frequencies. 
To investigate similar trends across fine-grained frequency band discretizations, 
we split the amplitude spectrum into $3$, $5$, $7$, and $9$ frequency bands in the manner described above, and analyze the diversity of these frequency bands across multiple datasets. Across $7$ domain shifts (see Fig.~\ref{fig:motivation_analysis_1} and~\ref{fig:motivation_analysis_2}) – $\{$GTAV, SYNTHIA$\}\to\{$Cityscapes, BDD100K, Mapillary$\}$, 
and VisDA-C Syn$\to$Real, we find that (1) for every dataset (whether synthetic or real), diversity decreases as we head towards higher frequency bands and (2) synthetic images exhibit less diversity in high-frequency bands at all considered levels of granularity. 

\par
\begin{figure*}
\centering
\includegraphics[width=\textwidth]{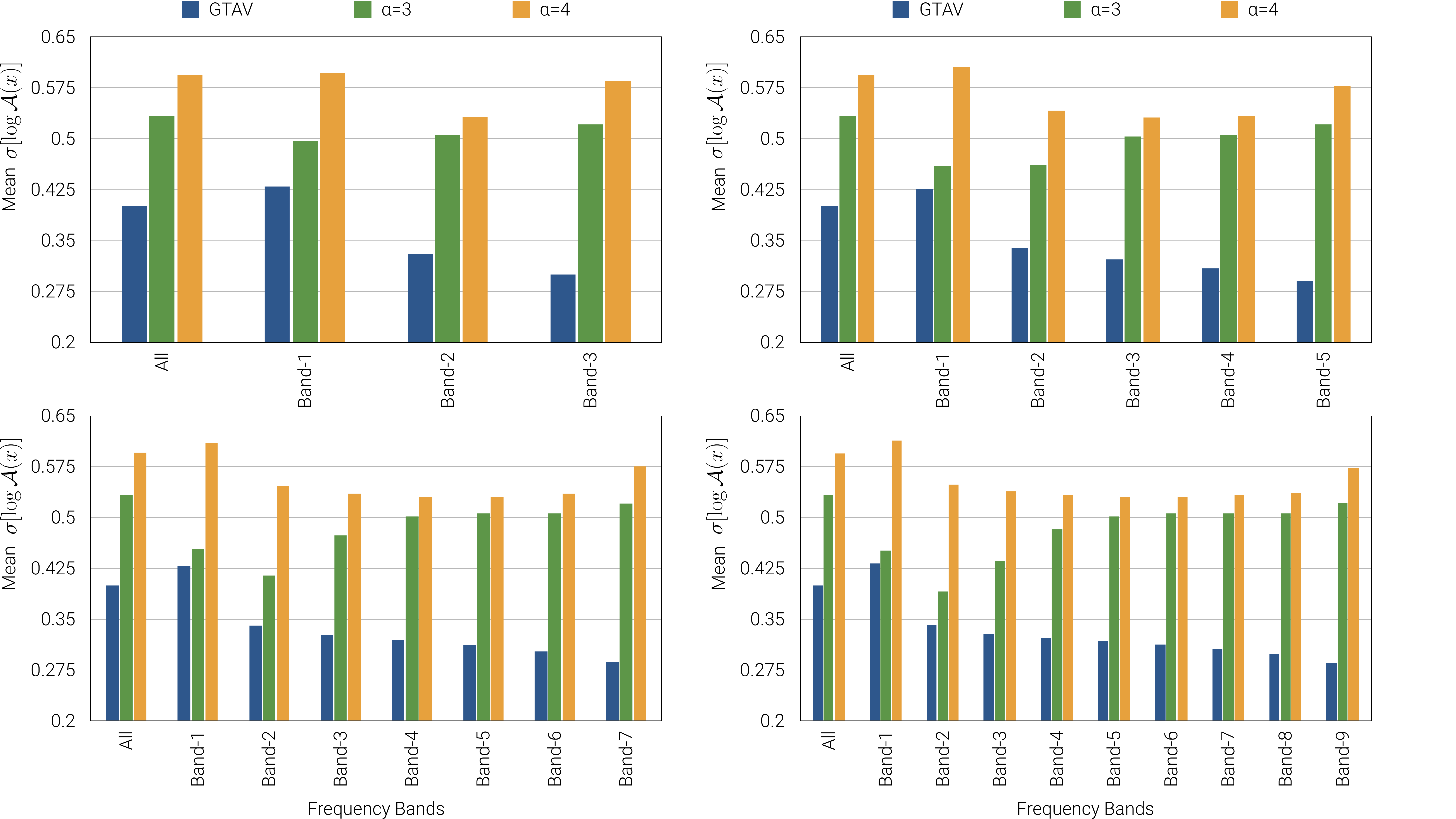}
\caption{\textbf{Variations in amplitude values across fine-grained frequency bands for synthetic images post-\pst (GTAV).} For GTAV, we find that applying \pst increases variations in amplitude values across different frequency bands. Four plots correspond to fine-grained frequency bands (3, 5, 7 and 9 bands; increasing in frequency from Band-1 to Band-$n$). We find the maximum amount of increase for the highest frequency bands across different granularity levels.}
\label{fig:pre_post_pasta_1}
\end{figure*}

\begin{figure*}
\centering
\includegraphics[width=\textwidth]{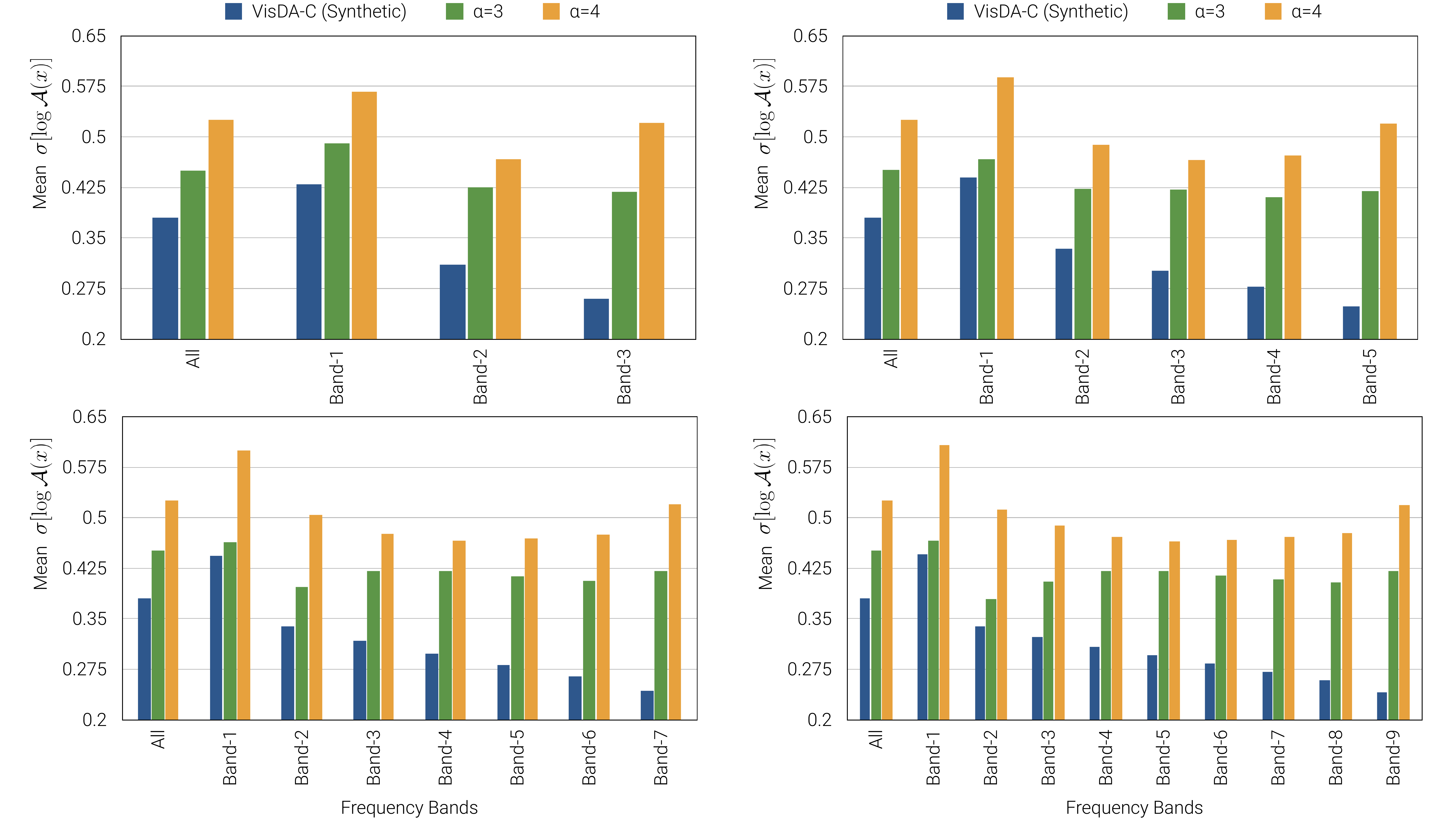}
\caption{\textbf{Variations in amplitude values across fine-grained frequency bands for synthetic images post-\pst (VisDA-C).} For VisDA-C (Synthetic), we find that applying \pst increases variations in amplitude values across different frequency bands. Four plots correspond to fine-grained frequency bands (3, 5, 7 and 9 bands; increasing in frequency from Band-1 to Band-$n$). We find the maximum amount of increase for the highest frequency bands across different granularity levels.}
\label{fig:pre_post_pasta_2}
\end{figure*}

 \noindent
\textbf{Increase in amplitude variations post-\pst.} Next, we observe how \pst effects the diversity of the amplitude spectrums on GTAV and VisDA-C. Similar to above, we split the amplitude spectrum into $3$, $5$, $7$, and $9$ frequency bands, and we analyze the diversity of these frequency bands before and after applying \pst to images (see Fig.~\ref{fig:pre_post_pasta_1} and~\ref{fig:pre_post_pasta_2}). For synthetic images from GTAV, when \pst 
is applied, we observe that the standard deviation of amplitude spectrums increases from $0.4$ to $0.497$, $0.33$ to $0.51$ and $0.3$ to $0.52$ for the low, mid and high frequency bands respectively. As expected, we observe maximum increase for the high-frequency bands.

%% file: appendix_sections/qualitative_examples.tex
\begin{figure*}
\centering
\includegraphics[width=\textwidth]{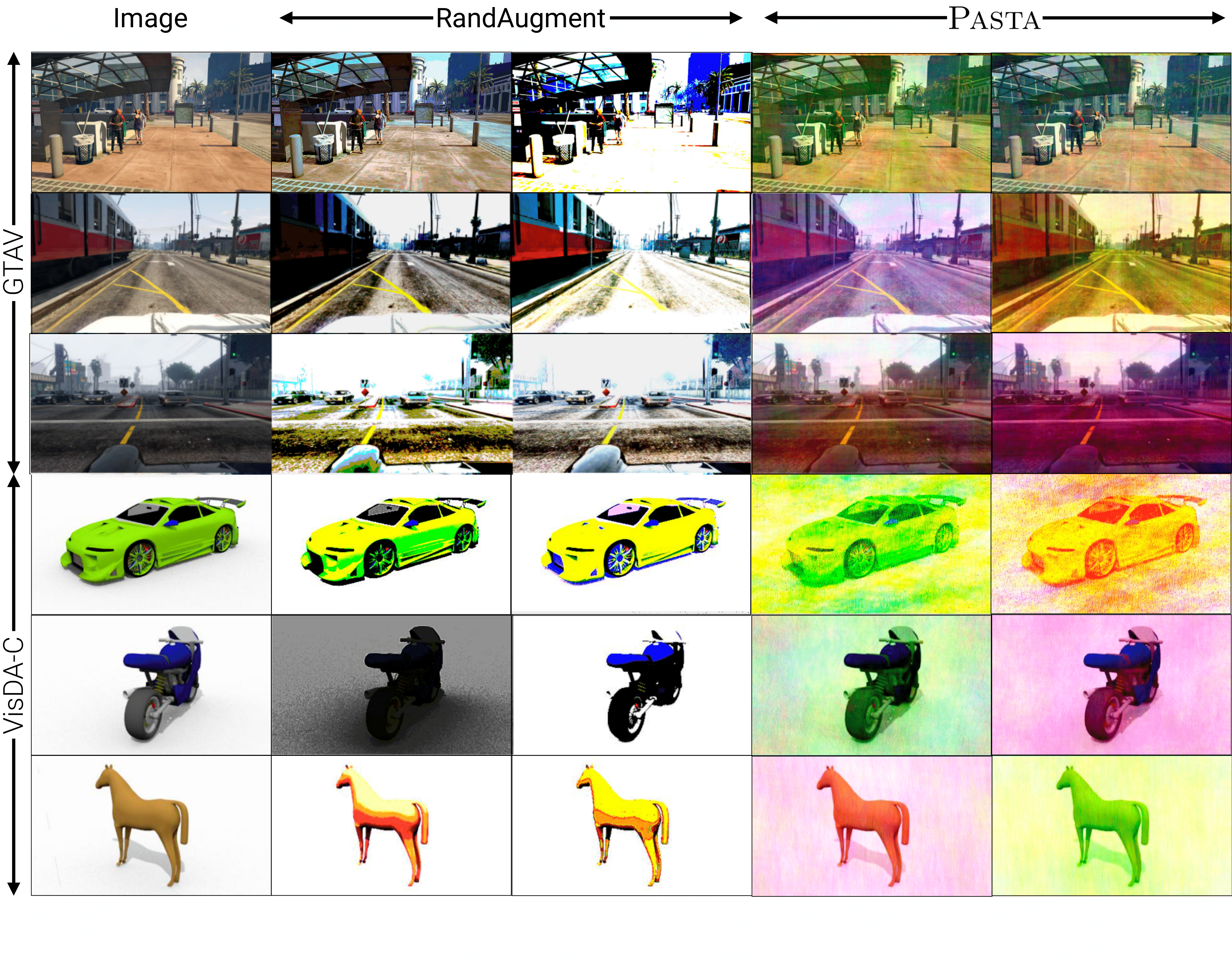}
\vspace{-10pt}
\caption{\textbf{\pst augmentation samples.} Examples of images from different synthetic datasets when augmented using \pst and RandAugment~\cite{cubuk2020randaugment}. Rows 1-3 include examples from GTAV and rows 4-6 from VisDA-C.
}
\label{fig:pasta_aug_samples}
\end{figure*}

\par \noindent
\textbf{\pst Augmentation Samples.} Fig.~\ref{fig:pasta_aug_samples} includes more examples of images from synthetic datasets (from GTAV and VisDA-C), when RandAugment and \pst are applied.

\begin{figure*}
\centering
\includegraphics[width=0.6\textwidth]{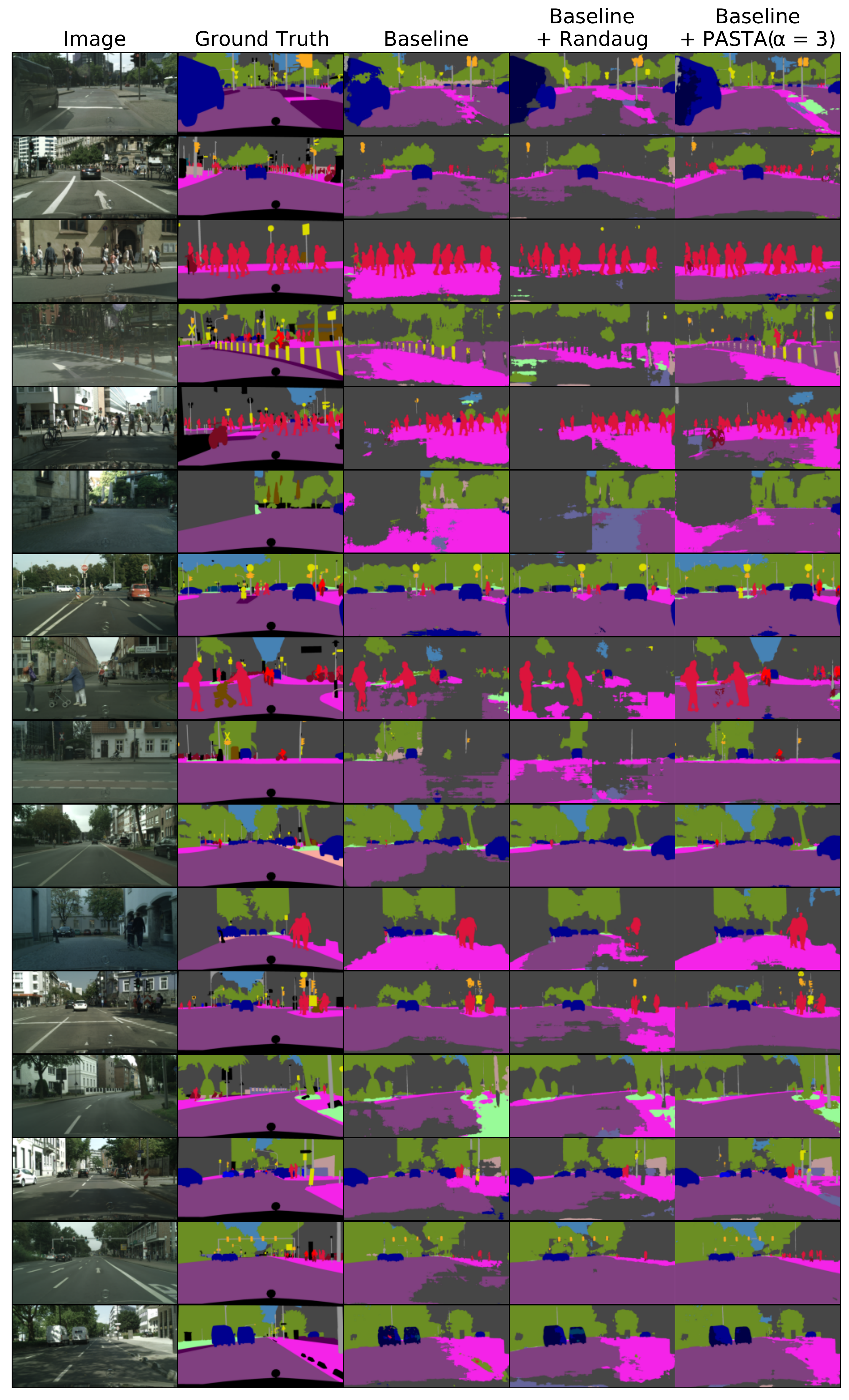}
\includegraphics[width=\textwidth]{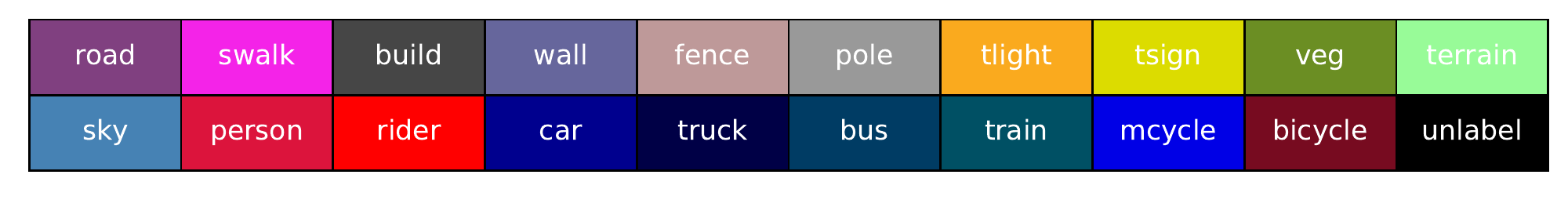}
\caption{\textbf{GTAV$\to$Cityscapes Baseline SemSeg Predictions.} Qualitative predictions made on randomly selected Cityscapes validation images by a Baseline DeepLabv3+ model (R-50 backbone) trained on GTAV synthetic images. The first two columns indicate the original image and the associated ground truth and rest indicate the listed methods.
}
\label{fig:base_qual_obs}
\end{figure*}

\begin{figure*}
\centering
\includegraphics[width=0.6\textwidth]{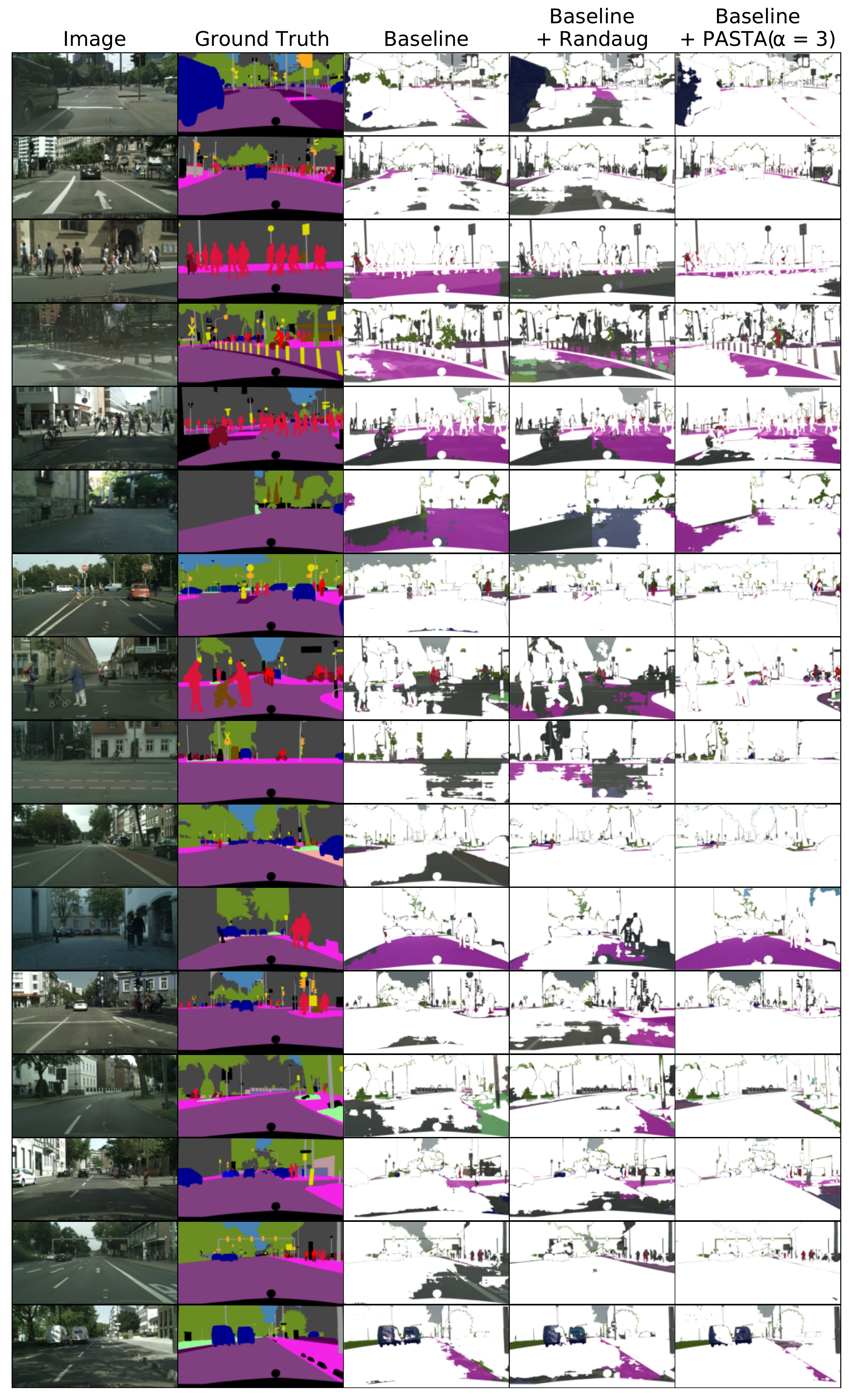}
\includegraphics[width=\textwidth]{appendix_figures/class_labels_ours.pdf}
\caption{\textbf{GTAV$\to$Cityscapes Baseline SemSeg Prediction Diffs.} Differences between prediction and ground truth for predictions made on randomly selected Cityscapes validation images by a Baseline DeepLabv3+ model (R-50 backbone) trained on GTAV synthetic images. The first two columns indicate the original image and the associated ground truth and rest indicate the listed methods.
}
\label{fig:base_diff_qual_obs}
\end{figure*}

\begin{figure*}
\centering
\includegraphics[width=0.5\textwidth]{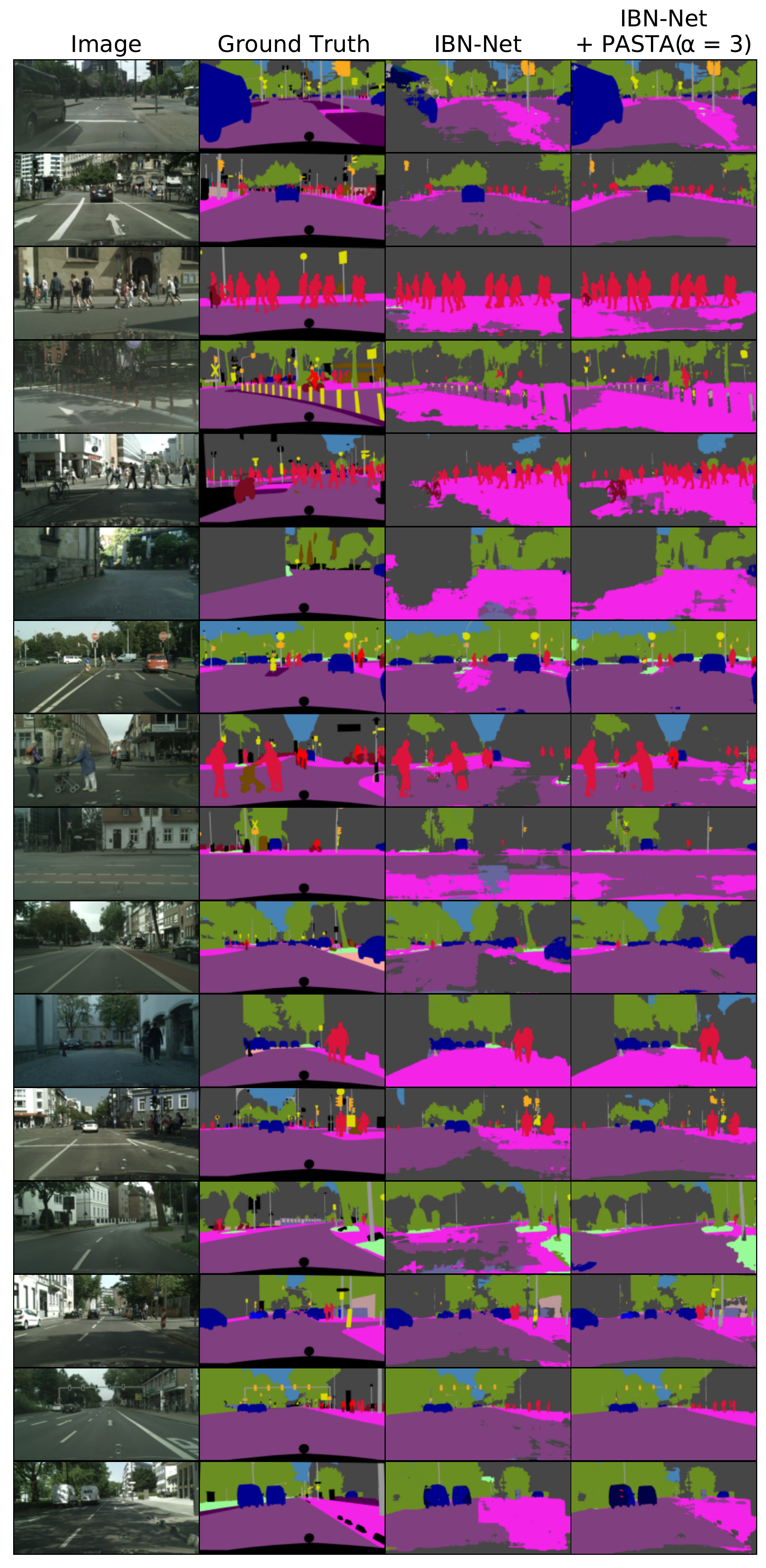}
\includegraphics[width=\textwidth]{appendix_figures/class_labels_ours.pdf}
\caption{\textbf{GTAV$\to$Cityscapes IBN-Net~\cite{pan2018two} SemSeg Predictions.} Qualitative predictions made on randomly selected Cityscapes validation images by IBN-Net (DeepLabv3+ model with R-50 backbone) trained on GTAV synthetic images. The first two columns indicate the original image and the associated ground truth and rest indicate the listed methods.
}
\label{fig:ibn_qual_obs}
\end{figure*}

\begin{figure*}
\centering
\includegraphics[width=0.5\textwidth]{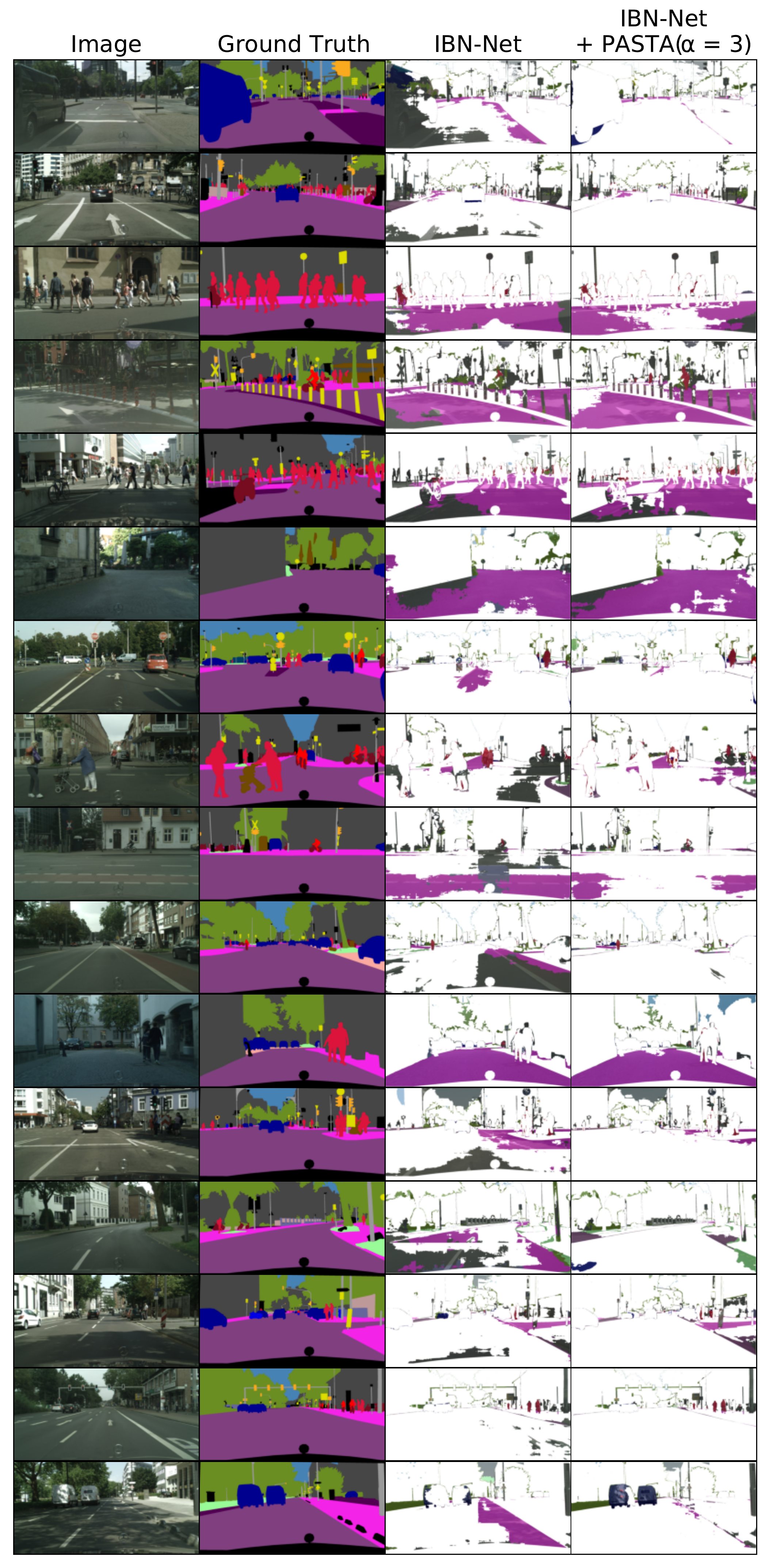}
\includegraphics[width=\textwidth]{appendix_figures/class_labels_ours.pdf}
\caption{\textbf{GTAV$\to$Cityscapes IBN-Net~\cite{pan2018two} SemSeg Prediction Diffs.} Differences between prediction and ground truth for predictions made on randomly selected Cityscapes validation images by IBN-Net (DeepLabv3+ model with R-50 backbone) trained on GTAV synthetic images. The first two columns indicate the original image and the associated ground truth and rest indicate the listed methods.
}
\label{fig:ibn_diff_qual_obs}
\end{figure*}

\begin{figure*}
\centering
\includegraphics[width=0.5\textwidth]{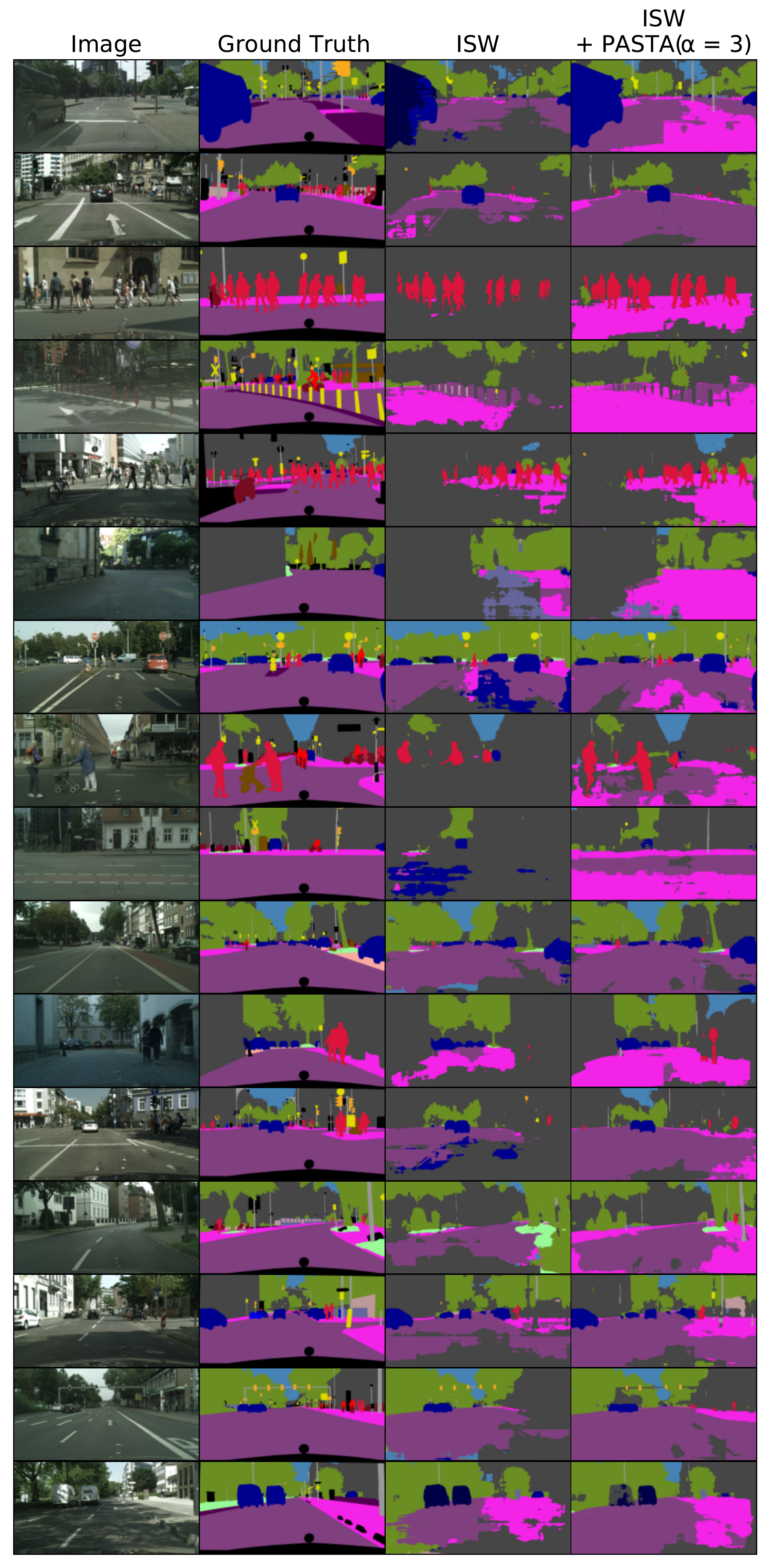}
\includegraphics[width=\textwidth]{appendix_figures/class_labels_ours.pdf}
\caption{\textbf{GTAV$\to$Cityscapes ISW~\cite{choi2021robustnet} SemSeg Predictions.} Qualitative predictions made on randomly selected Cityscapes validation images by ISW (DeepLabv3+ model with R-50 backbone) trained on GTAV synthetic images. The first two columns indicate the original image and the associated ground truth and rest indicate the listed methods.
}
\label{fig:isw_qual_obs}
\end{figure*}

\begin{figure*}
\centering
\includegraphics[width=0.5\textwidth]{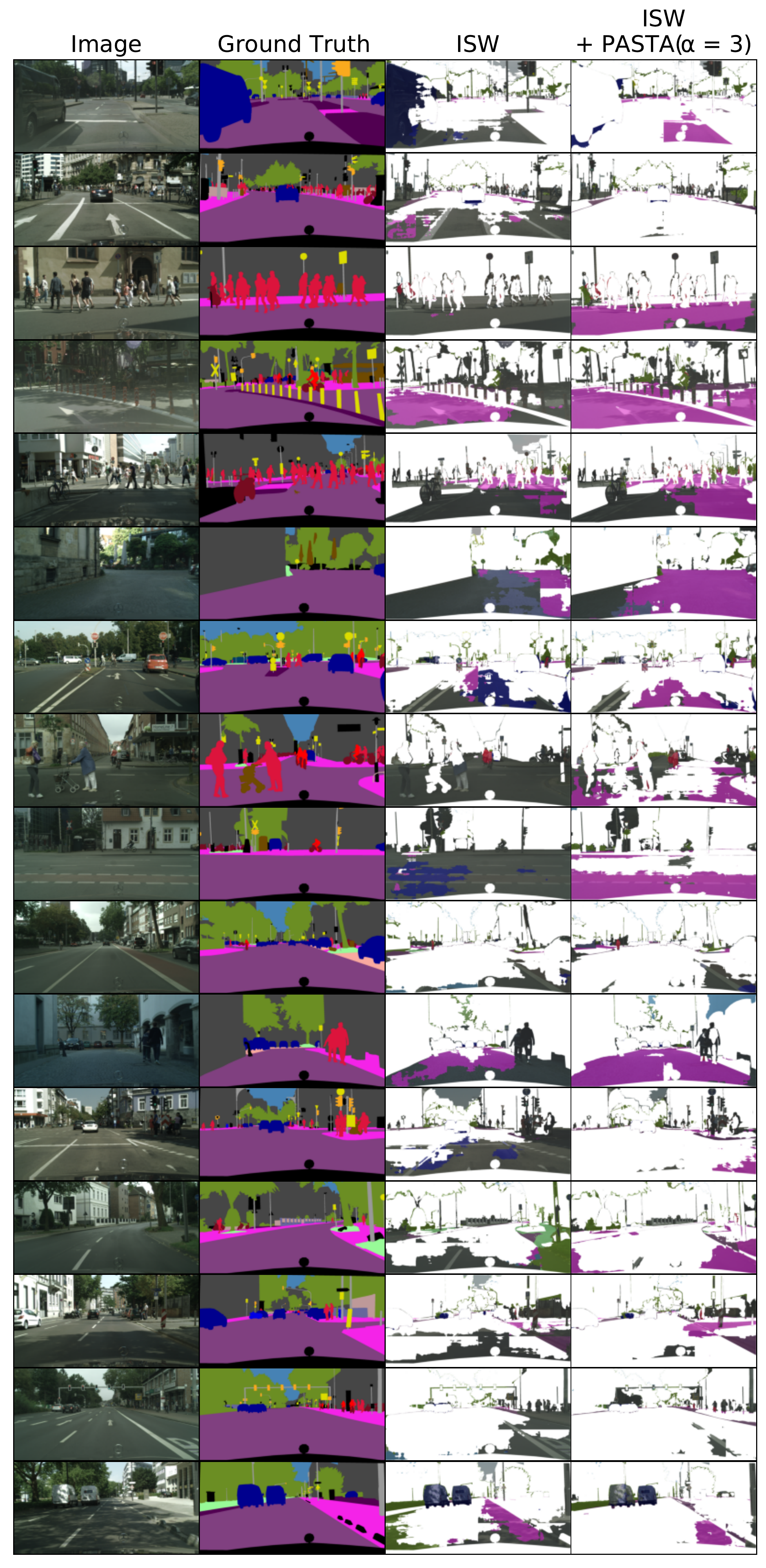}
\includegraphics[width=\textwidth]{appendix_figures/class_labels_ours.pdf}
\caption{\textbf{GTAV$\to$Cityscapes ISW~\cite{choi2021robustnet} SemSeg Prediction Diffs.} Differences between prediction and ground truth for predictions made on randomly selected Cityscapes validation images by ISW (DeepLabv3+ model with R-50 backbone) trained on GTAV synthetic images. The first two columns indicate the original image and the associated ground truth and rest indicate the listed methods.}
\label{fig:isw_diff_qual_obs}
\end{figure*}

\par \noindent
\textbf{Semantic Segmentation Predictions.} We include qualitative examples of semantic segmentation predictions on Cityscapes made by Baseline, IBN-Net and ISW (DeepLabv3+, ResNet-50) trained on GTAV (corresponding to 
Tables 1, 4 and 6 in the main paper)
Fig.~\ref{fig:base_qual_obs},~\ref{fig:ibn_qual_obs} and~\ref{fig:isw_qual_obs} respectively when different augmentations are applied (RandAugment and \pst). The Cityscapes images we show predictions on were selected randomly. We include RandAugment predictions only for the Baseline. To get a better sense of the kind of mistakes made by different approaches, we also include the difference between the predictions and ground truth segmentation masks in Fig.~\ref{fig:base_diff_qual_obs},~\ref{fig:ibn_diff_qual_obs} and~\ref{fig:isw_diff_qual_obs} (ordered accordingly for easy reference). The difference images show the predicted classes only for pixels where the prediction differs from the ground truth.

%% file: appendix_sections/asset_licenses.tex
The assets used in this work can be grouped into three categories -- Datasets, Code Repositories and Dependencies. We include the licenses of each of these assets below.

\par \noindent
\textbf{Datasets.} We used the following publicly available datasets in this work -- GTAV~\cite{richter2016playing}, Cityscapes~\cite{cordts2016cityscapes}, BDD100K~\cite{yu2020bdd100k}, Mapillary~\cite{neuhold2017mapillary}, Sim10K~\cite{johnson2016driving}, and VisDA-C~\cite{peng2017visda}. For GTAV, the codebase used to extract data from the original GTAV game is distributed under the MIT license.\footnote{\href{https://bitbucket.org/visinf/projects-2016-playing-for-data/src/master/}{https://bitbucket.org/visinf/projects-2016-playing-for-data/src/master/}} The license agreement for the Cityscapes dataset dictates that the dataset is made freely available to academic and non-academic entities for non-commercial purposes such as academic research, teaching, scientific publications, or personal experimentation and that permission to use the data is granted under certain conditions.\footnote{\href{https://www.cityscapes-dataset.com/license/}{https://www.cityscapes-dataset.com/license/}} BDD100K is distributed under the BSD-3-Clause license.\footnote{\href{https://github.com/bdd100k/bdd100k/blob/master/LICENSE}{https://github.com/bdd100k/bdd100k/blob/master/LICENSE}} Mapillary images are shared under a CC-BY-SA license, which in short means that anyone can look at and distribute the images, and even modify them a bit, as long as they give attribution.\footnote{\href{https://help.mapillary.com/hc/en-us/articles/115001770409-Licenses}{https://help.mapillary.com/hc/en-us/articles/115001770409-Licenses}} 
Densely annotated images for Sim10k are available freely\footnote{\href{https://fcav.engin.umich.edu/projects/driving-in-the-matrix}{https://fcav.engin.umich.edu/projects/driving-in-the-matrix}} and can only be used for non-commercial applications.
The VisDA-C development kit on github does not have a license associated with it, but it does include a Terms of Use, which primarily states that the dataset must be used for non-commercial and educational purposes only.\footnote{\href{https://github.com/VisionLearningGroup/taskcv-2017-public/tree/master/classification}{https://github.com/VisionLearningGroup/taskcv-2017-public/tree/master/classification}}

\par \noindent
\textbf{Code Repositories.} For our experiments, apart from code that we wrote ourselves, we build on top of three existing public repositories -- RobustNet\footnote{\href{https://github.com/shachoi/RobustNet}{https://github.com/shachoi/RobustNet}}, MMDetection\footnote{\href{https://github.com/open-mmlab/mmdetection}{https://github.com/open-mmlab/mmdetection}} and CSG\footnote{\href{https://github.com/NVlabs/CSG}{https://github.com/NVlabs/CSG}}. RobustNet is distributed under the BSD-3-Clause license. MMDetection is distributed under Apache License 2.0\footnote{\href{https://github.com/open-mmlab/mmdetection/blob/master/LICENSE}{https://github.com/open-mmlab/mmdetection/blob/master/LICENSE}}. CSG, released by NVIDIA, is released under a NVIDIA-specific license.\footnote{\href{https://github.com/NVlabs/CSG/blob/main/LICENSE.md}{https://github.com/NVlabs/CSG/blob/main/LICENSE.md}}

\par \noindent
\textbf{Dependencies.} We use Pytorch~\cite{NEURIPS2019_9015} as the deep-learning framework for all our experiments. Pytorch, released by Facebook, is distributed under a Facebook-specific license.\footnote{\href{https://github.com/pytorch/pytorch/blob/master/LICENSE}{https://github.com/pytorch/pytorch/blob/master/LICENSE}}